\documentclass{article}

\usepackage{microtype}
\usepackage{graphicx}
\usepackage{booktabs} 
\usepackage[ruled,linesnumbered]{algorithm2e}
\usepackage{multirow}
\usepackage{bbding}
\usepackage{colortbl}
\usepackage{lipsum}

\usepackage{hyperref}

\usepackage[accepted]{icml2024}

\usepackage{amsmath}
\usepackage{amssymb}
\usepackage{mathtools}
\usepackage{amsthm}
\usepackage{subfig}

\usepackage[capitalize,noabbrev]{cleveref}

\theoremstyle{plain}
\newtheorem{theorem}{Theorem}[section]

\theoremstyle{definition}

\newtheorem{assumption}[theorem]{Assumption}
\theoremstyle{remark}

\usepackage[textsize=tiny]{todonotes}

\begin{document}

\twocolumn[
\icmltitle{Recovering Labels from Local Updates in Federated Learning}



\icmlsetsymbol{equal}{*}

\begin{icmlauthorlist}
\icmlauthor{Huancheng Chen}{1}
\icmlauthor{Haris Vikalo}{1}
\end{icmlauthorlist}

\icmlaffiliation{1}{University of Texas at Austin, Texas, USA}

\icmlcorrespondingauthor{Huancheng Chen}{huanchengch@utexas.edu}
\icmlcorrespondingauthor{Haris Vikalo}{hvikalo@ece.utexas.edu}

\icmlkeywords{Federated Learning, Label Recovery}

\vskip 0.3in
]




\begin{abstract}
Gradient inversion (GI) attacks present a threat to the privacy of clients in federated learning (FL) by aiming to enable reconstruction of the clients' data from communicated model updates. A number of such techniques attempts to accelerate data recovery by first reconstructing labels of the samples used in local training. 
However, existing label extraction methods make strong assumptions that typically do not hold in realistic FL settings. In this paper we present a novel label recovery scheme, {\it 
\underline{R}ecovering \underline{L}abels from Local \underline{U}pdates}
(RLU), which provides near-perfect accuracy when attacking untrained (most vulnerable) models. More significantly, RLU achieves high performance even in realistic real-world settings where the clients in an FL system run multiple local epochs, train on heterogeneous data, and deploy various optimizers to minimize different objective functions. Specifically, RLU estimates labels by solving a least-square problem that emerges from the analysis of the correlation between labels of the data points used in a training round and the resulting update of the output layer. The experimental results on several datasets, architectures, and data heterogeneity scenarios demonstrate that the proposed method consistently outperforms existing baselines, and helps improve quality of the reconstructed images in GI attacks in terms of both PSNR and LPIPS.
\end{abstract}

\section{Introduction}
\label{introduction}

Federated learning (FL) \citep{fedavg}, which aims to enable collaborative training of ML models while protecting privacy of the participating clients, has attracted considerable interest in privacy-sensitive fields such as healthcare and finance \citep{yangqiang}. However, recent works have demonstrated vulnerability of FL systems to privacy attacks such as membership inference \citep{membership}, property inference \citep{property} and gradient inversion \citep{dlg}. In particular, gradient inversion methods have been shown capable of reconstructing private training data given the gradients computed during local training. The milestone work, Deep Leakage from Gradient (DLG) \citep{dlg}, minimizes the difference between simulated and true gradients to extract data and the corresponding labels belonging to FL clients training a model for a classification task. 
However, DLG does not perform well in settings where the training batch size is large and the data resolution is high, since in such scenarios the joint optimization of data and the corresponding labels becomes challenging. A subsequent study, iDLG \citep{idlg}, presents an analytical method for the recovery of ground-truth labels from the gradients by exploiting a relationship between the labels and the signs of the gradients. 
GradInversion \citep{yin} takes a step further by exploiting a relationship between labels and the magnitudes of gradient in the output layer to perfectly restore labels of the samples used in the considered training round, and facilitates high-resolution reconstruction of data in a batch with as many as 48 samples. Unfortunately, this method assumes no repeated labels in the batch, which is unrealistic in real-world settings. Recently, iRLG \citep{irlg} attempted to address the limitations of GradInversion, enabling recovery of potentially repeated labels in a batch by utilizing gradients computed using a randomly initialized untrained model. However, the performance of iRLG drastically deteriorates as the model's accuracy improves. 

The above label recovery methods either: (1) assume no repeated labels in a batch; (2) use non-negative network activation functions; or (3) perform well only on untrained models. They further assume that each client in an FL system runs only a single epoch when updating its local model, while the more practical settings with multiple epochs of local training are not considered. Moreover, these studies are limited to the scenarios where a server collects gradients
and do not apply to the standard FL setting where
the server collects model updates rather than the gradients.

Aiming to address the aforementioned limitations of the prior work, in this paper we propose a novel method for label recovery, \underline{R}ecovering \underline{L}abels from Local \underline{U}pdates (RLU). We start by analyzing correlations between local updates of the output layer and the ground-truth labels of data samples in training batches for several frequently used FL algorithms. These correlations, along with the expected value of the ``erroneous confidence" (i.e., the level of confidence when making erroneous decisions) evaluated on a small auxiliary dataset, are then used to recover the labels. As local training unfolds across multiple epochs, distribution of the previously mentioned ``erroneous confidence" undergoes changes which the server cannot access. Simulating an underlying dynamical model helps estimate how this distribution evolves across training epochs; the obtained estimates of intermediate distributions are then utilized to help achieve notable enhancement in the accuracy of label recovery. The effectiveness of RLU is demonstrated in extensive experiments involving several FL algorithms and various model architectures on SVHN, CIFAR10, CIFAR100 and Tiny-ImageNet datasets. In all the considered settings, RLU outperforms state-of-the-art baselines on both untrained and well-trained models. The main contributions of the paper are summarized as follows:
\begin{itemize}
\item We propose a general analytical method that enables the server in an FL system to recover labels of the points used by clients in local training; the proposed method applies to various FL algorithms and makes no assumptions regarding the activation function or batch label composition.

\item To allow accurate label recovery in FL systems where clients train across multiple epochs, we simulate the evolution of the model through the epochs via a Monte Carlo method that updates the ``erroneous confidence" characterizing the correlation between local updates and training data labels.

\item To evaluate the proposed label recovery method, we conduct comprehensive benchmarking experiments across a number of FL settings where we vary the level of data heterogeneity, model architectures and local objective functions. 
\end{itemize}

\section{Preliminary and Related Work}
\label{related_work}
\textbf{Federated Learning}. The classical FL algorithm, FedAvg \citep{fedavg}, enables $K$ clients to collaboratively train global model without sharing their private data $\mathcal{D}_{k}$. In FedAvg, the server initializes training round $t$ by broadcasting global model $\boldsymbol{\theta}^{(t)}$ to the clients. Client $k$ updates its local model by running a gradient descent procedure on its (private) local data; the server then collects updated local models $\boldsymbol{\theta}^{(t)}_{k}$ from the clients and averages them to form a new global model,
\vspace{-0.1in}
\begin{equation}
    \boldsymbol{\theta}^{(t+1)} = \sum_{k=1}^{K}p_{k}\boldsymbol{\theta}^{(t)}_{k},
\end{equation}
where $p_{k}$ denotes the weight assigned to client $k$. Since the training is initialized by broadcasting global model $\boldsymbol{\theta}^{(t)}$ to the clients, local updates are computed as $\Delta\boldsymbol{\theta}^{(t)}_{k} = \boldsymbol{\theta}^{(t)}_{k} - \boldsymbol{\theta}^{(t)}$. It is worth pointing out that in FL algorithms other than FedAvg, $\Delta\boldsymbol{\theta}^{(t)}_{k}$ is not necessarily proportional to the gradients $\nabla \boldsymbol{\theta}^{(t)}_{k}$. We discuss properties of local updates in Section \ref{beyond_fedavg}.

\textbf{Gradients Inversion Attack}. DLG \citep{dlg} is the first optimization-based method for reconstructing training data given the gradients $\nabla \boldsymbol{\theta}$ and model $\boldsymbol{\theta}$. More specifically, DLG minimizes the difference between simulated and ground-truth gradients,
\begin{equation}
\label{ig}
    \mathbf{x}^{\star}, \mathbf{y}^{\star} = \text{arg}\min_{\mathbf{x}^{\prime}, \mathbf{y}^{\prime}} \left\Vert \nabla_{\boldsymbol{\theta}} \mathcal{L}_{\text{task}}(\boldsymbol{\theta}, \mathbf{x}^{\prime}, \mathbf{y}^{\prime}) - \nabla \boldsymbol{\theta} \right\Vert^{2}.
\end{equation}
Follow-up studies \citep{geiping, yin} introduced total variation and group consistency regularization to the objective of the gradient inversion optimization, enabling high reconstruction performance on ImageNet \citep{imagenet}. Recently, a number of works \citep{generative,gifd,overparameter} leveraged pre-trained generative models to improve the gradient inversion attack and achieve state-of-the-art performance with batch size set to $32$. However, these schemes assume knowing the ground-truth labels $\mathbf{y}$ in the batch -- an unrealistic assumption that significantly accelerates the search for the optimal data $\mathbf{x}^{\star}$.

\textbf{Label Recovery Attack}. To improve the performance of an attack on gradients computed using samples coming from a relatively large batch, a series of studies \citep{idlg,yin,rlg,llg,zlg, irlg} proposed various analytical approaches to recovering labels $\mathbf{y}$ prior to solving the optimization over $\mathbf{x}^{\prime}$. However, all these methods suffer limitations that restrict their practical feasibility. GradInversion \citep{yin} assumes there are no repeated labels in a batch; RLG \citep{rlg} can only recover class-wise labels but not instance-wise labels; LLG \citep{llg} and ZLG \citep{zlg} require non-negative activation functions; iRLG \citep{irlg} performs well only on randomly initialized untrained models. 
Moreover, these methods primarily focus on FedAvg and generally provide little if any discussion of other FL algorithms.

\vspace{-0.1 in}
\section{Methodology}
\label{methodology}
\subsection{Problem Settings}
\label{problem_formulation}
We consider multi-class classification models trained in FL settings where an honest-but-curious (HBC) server aims to recover ground-truth labels of the training samples using local updates $\Delta \boldsymbol{\theta}^{(t)}_{k}$ collected from clients while following a standard FL training procedure. The HBC server knows the batch size, number of local epochs and learning rate used by the clients but has no information about local data distribution. We assume each local model is trained by minimizing the cross-entropy (CE) loss 
\begin{equation}
\label{celoss}
    \mathcal{L}_{\text{ce}} = -\frac{1}{|\mathcal{B}|}\sum_{i = 1}^{|\mathcal{B}|}\log\frac{\exp(\mathbf{q}_{y^{(i)}}^{(i)})}{\sum_{n = 1}^{N} \exp(\mathbf{q}_{n}^{(i)})}, 
\end{equation}
where $\mathcal{B}$ denotes a training batch; $(\mathbf{x}^{(i)}, y^{(i)})$ is the $i$-th example in the batch; $N$ is the number of classes; and $\mathbf{q}^{(i)} = \mathbf{W} \cdot \mathbf{e}^{(i)} + \mathbf{b}$ denotes the output logits of the model given the embedding $\mathbf{e}^{(i)} \in \mathbb{R}^{L}$ of $\mathbf{x}^{(i)}$, where $\mathbf{W} \in \mathbb{R}^{N\times L}$ and $\mathbf{b} \in \mathbb{R}^{N}$ are the weights and bias of the output layer.

\subsection{ Label Recovery from Local Updates (RLU)}
\label{charateristic}
Following the definition of the CE loss in Eq. \ref{celoss}, contribution of $\mathbf{x}^{(i)}$ to the $j$-th component of the gradient of $\mathbf{b}$ can be computed as (the proof provided in Appendix \ref{gradient_bias})
\begin{equation}
    \nabla\mathbf{b}_{j}^{(i)} = \left\{
    \begin{aligned}
        & \;\;\;\; \frac{\exp(\mathbf{q}_{j}^{(i)})}{\sum_{n = 1}^{N} \exp(\mathbf{q}_{n}^{(i)})} = \mathbf{s}_{j}(\mathbf{x}^{(i)}), \text{ if } j \not = y^{(i)}, \\
        & -\frac{\sum_{n \not = j}\exp(\mathbf{q}_{n}^{(i)})}{\sum_{n = 1}^{N} \exp(\mathbf{q}_{n}^{(i)})}, \text{ if } j  = y^{(i)}.\\
    \end{aligned}
    \right.
\end{equation}
Assuming stochastic gradient descent (SGD) optimizer, the local update of the $j$-th component of the output layer's bias computed by client $k$ is 
\begin{equation}
\label{local_update}
     \Delta\mathbf{b}_{j} = -\frac{\eta}{|\mathcal{B}^{\tau}|}\sum_{\tau = 1}^{m}\sum_{i=1}^{|\mathcal{B}|}\nabla\mathbf{b}_{j}^{(i,\tau)}, \mathcal{B}^{\tau}\sim \mathcal{D}_{k}, 
\end{equation}
where $\tau$ is the local epoch index, $m$ is the number of epochs, $\eta$ is the learning rate, and $\mathcal{D}_{k}$ denotes client $k$'s data. Note that $\mathbf{s}_{j}(\mathbf{x}^{(i)}) \in (0,1)$ can be interpreted as the ``erroneous confidence" of labeling $\mathbf{x}^{(i)}$ as class $j$ while $j \not = y^{(i)}$ (i.e., confidence in a labeling decision that is in fact erroneous). Let $\mathcal{S}_{n,j}$ denote the expected erroneous confidence for class $j$ given a sample with true label $n \not = j$,
\begin{equation}
    \mathcal{S}_{n,j} = \mathbb{E}_{(\mathbf{x}, y)\sim \mathcal{D}_{k}^{(n)}}\left[ \mathbf{s}_{j}(\mathbf{x})\right], \forall n,j \in [N] \wedge n \not = j,
\end{equation}
where $\mathcal{D}_{k}^{(n)} \subseteq \mathcal{D}_{k}$ collects samples with label $n$ and $(\mathbf{x}, y)$ is a random sample from $\mathcal{D}_{k}^{(n)}$. These expectations are indicative of the model's training error: $\mathcal{S}_{n,j} \approx \frac{1}{N}$ in a randomly initialized untrained model making random predictions, whereas $\mathcal{S}_{n,j}$ asymptotically goes to $0$ as the accuracy of the model increases.

Note that the expected erroneous confidence, $\mathcal{S}_{n,j}$ does not admit closed-form expression; to analyze it and gain needed insight, we make the following assumption.
\begin{assumption} 
\label{assumption_1}
The output logits $\mathbf{q}^{(n)}$ of model $\boldsymbol{\theta}$ when the input is $\mathbf{x} \sim \mathcal{D}^{(n)}$ follows a multivariate normal distribution, i.e.,
\begin{equation}
    \mathbf{q}^{(n)} \sim \mathcal{N}(\boldsymbol{\mu}_{n}, \boldsymbol{\Sigma}_{n}),
\label{assume31}
\end{equation}
where the mean $\boldsymbol{\mu}_{n}$ and covariance $\boldsymbol{\Sigma}_{n}$ depend on the accuracy of the model. 
\end{assumption}

For an untrained deep model whose parameters are initialized from a zero-mean uniform distribution, expected values of the output logits $\boldsymbol{\mu}_{1}, \dots, \boldsymbol{\mu}_{N}$ are approximately $\mathbf{0}$. As the accuracy of the model improves during the training process, $\boldsymbol{\mu}_{j,j}$ converges to a positive value while $\boldsymbol{\mu}_{n,j} (n \not= j)$ converges to a negative value, forcing $\mathcal{S}_{n,j}$ to converge to $0$. The existing methods that perform well only on untrained models operate under the assumption $\boldsymbol{\mu}_{n} = \mathbf{0}$, which does not hold for well-trained models. Experimental results that empirically verify Assumption~3.1 are provided in Appendix~\ref{validation_assumption1}.

In each global round of training, the server leverages global model $\boldsymbol{\theta}^{(t)}$ to obtain estimates of the parameters in (\ref{assume31}), $\bar{\boldsymbol{\mu}}_{n}$ and $\bar{\boldsymbol{\Sigma}}_{n}$, via a Monte Carlo method run on a small auxiliary dataset $\mathcal{A}$: the samples from $\mathcal{A}$ are processed by $\boldsymbol{\theta}^{(t)}$ and the resulting output logits are used to empirically compute the mean and variance. Given the estimates of the parameters, the server samples $M$ data points $ \mathbf{q}^{(n,i)} \sim \mathcal{N}(\bar{\boldsymbol{\mu}}_{n}, \bar{\boldsymbol{\Sigma}}_{n})$ to infer $\mathcal{S}_{n,j}$ as
\begin{equation}
\label{monte_carlo}
    \bar{\mathcal{S}}_{n,j} = \frac{1}{M}\sum_{i=1}^{M}\frac{\exp(\mathbf{q}^{(n,i)}_{j})}{\sum_{c=1}^{N}\exp (\mathbf{q}_{c}^{(n,i)})}.
\end{equation}

\subsubsection{Single Epoch Local Training}
When $m = 1$, we omit superscript $\tau$ and let
$\Delta\mathbf{b}_{j} = -\frac{\eta}{|\mathcal{B}|}\sum_{i=1}^{|\mathcal{B}|}\nabla\mathbf{b}_{j}^{(i)}$. By taking the expectation of $\Delta\mathbf{b}_{j}$, we obtain (details provided in Appendix~\ref{expectation_bias_appendix})
\begin{equation}
\label{eq_b}
    \mathbb{E}\left[\Delta\mathbf{b}_{j}\right] = \frac{\eta}{|\mathcal{B}|}\left(N_{j}\sum_{n \not= j} \mathcal{S}_{j,n} - \sum_{n \not = j} N_{n}\mathcal{S}_{n,j}\right),
\end{equation}
where $N_{j}$ denotes the number of samples in $\mathcal{B}$ with label $j$. Finally, $N_{j}$ is estimated by solving  
\begin{equation}
\label{minimization}
\begin{aligned}
& \min_{\mathbf{z}\in \mathbb{R}^{N}} \left\Vert \mathbf{A}\mathbf{z} -  \mathbf{u}\right\Vert^{2}_{2} \\
\text{s.t.}\quad & \mathbf{0} \leq \mathbf{z} \leq  \mathbf{1} \wedge \left\Vert\mathbf{z}\right\Vert_{1} = 1,
\end{aligned}
\end{equation}
where $\mathbf{u} = \Delta \mathbf{b}/\eta$, the diagonal entries of the coefficient matrix $\mathbf{A}$ are $a_{j,j} = \sum_{n\not = j} \mathcal{S}_{j,n}$, and the $(n,j)$ off-diagonal entry of $\mathbf{A}$ is $a_{n,j} = -\mathcal{S}_{n,j}$. After finding the solution $\mathbf{z}^{\star}$ to the above problem, we estimate $N_{j}$ as $\bar{N}_{j} = \lfloor|\mathcal{B}|\cdot \mathbf{z}^{\star}_{j}\rceil$.

\subsubsection{Multiple Epochs of Local Training}
\label{multiple_epochs}
To reduce communication bandwidth, clients in FL typically update their local models over multiple epochs (with multiple batches of data), as illustrated in Eq.~\ref{local_update}. Assume fixed learning rate $\eta$ and constant batch size across all epochs (i.e., $|\mathcal{B}^{\tau}| = |\mathcal{B}|$ for all $\tau$); then the expectation of the local update $\mathbb{E}\left[\Delta\mathbf{b}_{j}^{(t)}\right]$ in global round $t$ can be found as
\begin{equation}
\label{eq_b_multi}
      \frac{\eta}{|\mathcal{B}|}\sum_{\tau = 1}^{m}\left(N_{j}^{(t,\tau)}\sum_{n \not = j} \mathcal{S}_{j,n}^{(t,\tau)} - \sum_{n \not = j} N_{n}^{(t,\tau)}\mathcal{S}_{n,j}^{(t,\tau)}\right),
\end{equation}
where $N_{j}^{(t,\tau)}$ is the number of samples with label $j$ in epoch $\tau$ and $\mathcal{S}_{n,j}^{(t,\tau)}$ denotes the expected erroneous confidence of local model $\boldsymbol{\theta}_{k}^{(t,\tau-1)}$ on class $j$. At the beginning of local training, $\boldsymbol{\theta}^{(t,0)}_{k}$ is initialized with the global model $\boldsymbol{\theta}^{(t)}$. Note that $\mathcal{S}_{n,j}^{(t,1)}$ can be estimated using global model $\boldsymbol{\theta}^{(t)}$ while $\mathcal{S}_{n,j}^{(t,m+1)}$ is readily inferred using the collected local model $\boldsymbol{\theta}^{(t,m)}_{k}$ via previously described Monte Carlo procedure. However, the intermediate states $\mathcal{S}_{n,j}^{(t,2)},\dots, \mathcal{S}_{n,j}^{(t,m)}$ are unknown since the server does not have access to local models $\boldsymbol{\theta}_{k}^{(t,1)},\dots,\boldsymbol{\theta}_{k}^{(t,m-1)}$. Since the evolution of $\mathcal{S}_{n,j}^{(t,\tau)}$ is 
non-linear, trivial interpolation between $\mathcal{S}_{j,n}^{(t,1)}$ and $\mathcal{S}_{j,n}^{(t,m+1)}$ may be highly inaccurate. To this end, we propose a practical method that relies on dynamics of the model parameter updates to approximate $\mathcal{S}_{j,n}^{(t,\tau)}$ in the intermediate epochs. Suppose we know $\mathcal{S}_{j,n}^{(t,\tau)}$ and the numbers of samples with different labels 
$N_{j}^{(t,\tau)}$, $j \in [N]$. Then 
\begin{equation}
\label{update_of_bias}
    \mathbb{E}\left[\Delta \mathbf{b}_{j}^{(t,\tau)}\right] = \frac{\eta}{|\mathcal{B}|}\left(N_{j}^{(t,\tau)}\sum_{n \not= j} \mathcal{S}_{j,n}^{(t,\tau)} - \sum_{n \not = j} N_{n}^{(t,\tau)}\mathcal{S}_{n,j}^{(t,\tau)}\right).
\end{equation}
Since the gradients of the weights in the output layer satisfy $\nabla  \mathbf{W}_{j,l} = \nabla  \mathbf{b}_{j} \cdot \bar{\mathbf{e}}_{l}$, where $\bar{\mathbf{e}}_{l}$ is the $l$-th component of the average embedded signal, we can estimate the change of the output logit as (the proof is provided in Appendix \ref{gradient_weight})
\begin{equation}
\label{use_signal_to_update}
    \mathbb{E}\left[\Delta \mathbf{q}_{j}^{(n)}\right] = \Delta \boldsymbol{\mu}_{n,j}^{(t,\tau)} = \mathbb{E}\left[\Delta \mathbf{b}_{j}^{(t,\tau)}\right]  \cdot \sum_{l=1}^{L}\bar{\mathbf{e}}_{l}^{2},
\end{equation}
and then obtain $\boldsymbol{\mu}_{n,j}^{(t,\tau+1)} = \boldsymbol{\mu}_{n,j}^{(t,\tau)} + \Delta\boldsymbol{\mu}_{n,j}^{(t,\tau)}$. Using $\boldsymbol{\mu}_{n,j}^{(t,\tau+1)}$ in the next local epoch, 
one can estimate $\mathcal{S}_{n,j}^{(t,\tau+1)}$ according to Eq. \ref{monte_carlo}. By recursively conducting the above procedure, one can estimate all the intermediate states $\mathcal{S}_{n,j}^{(t,\tau)}$. However, $N_{j}^{(t,\tau)}$ and $\bar{\mathbf{e}}_{l}$ are not known -- only the average updates of weight $\Delta \mathbf{W}_{j,l}^{(t)}$ and bias $\Delta \mathbf{b}_{j}^{t}$ are given. A closer examination of the correlation between $\nabla  \mathbf{W}_{j,l}^{(t)}$ and $\nabla \mathbf{b}_{j}^{t}$ suggests estimating the average embedded
signal according to
\begin{equation}
\label{expectation_of_bias_multiple}
    \bar{\mathbf{e}}_{l} \approx \Delta \mathbf{W}_{j,l}^{(t)} / \Delta \mathbf{b}_{j}^{(t)}.
\end{equation}
To estimate the total number $\{\bar{N}_{j}^{(t)}\}_{j=1}^{N}$ of labels in $m$ sampled batches we first set $N_{j}^{(t,\tau)} = \mathbf{g}_{j}$, where $\mathbf{g}\in \mathbb{N}^{N}$ denotes an arbitrarily vector (a guess) satisfying $\left\Vert \mathbf{g}\right\Vert_{1} = |\mathcal{B}|$. As described earlier in this subsection, if we knew $N_{j}^{(t,\tau)}$ we could dynamically update $\mathcal{S}_{j,n}^{(t,\tau)}$ to arrive at $\bar{\mathcal{S}}_{j,n}^{(t,m+1)}$. Since the true $\mathcal{S}_{j,n}^{(t,m+1)}$ is known by the server (collected after the final epoch), the difference between $\bar{\mathcal{S}}_{j,n}^{(t,m+1)}$ and $\mathcal{S}_{j,n}^{(t,m+1)}$ could be used to adjust $\mathbf{g}_{j}$ and subsequently improve the estimate $\bar{\mathcal{S}}_{j,n}^{(t,m+1)}$. If $\bar{\mathcal{S}}_{j,n}^{(t,m+1)}$ is significantly smaller than $\mathcal{S}_{j,n}^{(t,m+1)}$, $\mathbf{g}_{j}$ was underestimated; otherwise, $\mathbf{g}_{j}$ was overestimated. Intuitively, local model tends to label input data as class $j$ if the samples with label $j$ are dominant in the batches sampled for training.

After a number of iterations, the difference between $\bar{\mathcal{S}}_{j,n}^{(t,m+1)}$ and $\mathcal{S}_{j,n}^{(t,m+1)}$ becomes small and $m \cdot \mathbf{g}_{j}$ closely approximates $N_{j}^{(t)}$. To accelerate the search for $\mathbf{g}$, one can introduce $\bar{\mathcal{S}}_{j,n}^{(t)} = (\mathcal{S}_{j,n}^{(t,1)} + \mathcal{S}_{j,n}^{(t,m+1)})/2$ and solve optimization (\ref{minimization}) parameterized by $\mathcal{S}_{j,n} = \bar{\mathcal{S}}_{j,n}^{(t)}$ to obtain an initial estimate $\bar{N}_{j}^{(t)}$ which, in turn, is used to initialize $\mathbf{g}_{j} = \bar{N}_{j}^{(t)}/m$. In our experiments, following such an initialization RLU achieves highly accurate performance after only $T=5$ iterations. The algorithms described in this section are formalized in Appendix \ref{rlu} and \ref{search}.






\vspace{-0.1 in}
\subsection{FL schemes beyond FedAvg and SGD}
\label{beyond_fedavg}
\begin{table*}[t]
\caption{Coefficients $\rho^{(\tau)}$ and $\mathbf{h}_{j}^{(\tau)}$ for different FL schemes including FedAvg \citep{fedavg}, Scaffold \citep{scaffold}, FedProx \citep{fedprox}, FedDyn \citep{feddyn} and FedDC \citep{feddc}. Here $m$ denotes the number of local epochs, $\tau$ is the index of a local epoch, $\eta$ is the learning rate, and $\Delta \mathbf{b}_{j}^{(r)}$ denotes the $j$-th component of the local update of bias $\mathbf{b}$ in global round $r$. For FedDC, $\Delta \mathbf{B}_{j}^{(r)}$ denotes the $j$-th component of the global update of bias $\mathbf{b}$ while $\mathbf{d}_{k}^{(t)}$ is the local drift in global round $t$. The server in each of the schemes can collect the corresponding variables and potentially use them in a label recovery attack.} 
\centering
\small
\begin{tabular}{c|c|c|c|c}
\bottomrule[1pt]
\label{table1}
 Schemes    & Optimizer &$\rho^{(\tau)}$  & $\mathbf{h}_{j}^{(t)}$ & Regularizer  \\
 \hline
\multirow{5}{*}{FedAvg } & SGD      & $1$ & $\mathbf{0}$   &  No            \\   
\cline{2-5}
&  \multirow{2}{*}{SGDm} & \multirow{2}{*}{$\frac{1-\gamma^{m + 1 - \tau}}{1-\gamma}$}  & \multirow{2}{*}{$\mathbf{0}$} & \multirow{2}{*}{No; $\gamma$ is the momentum weight} \\
         &  &                &     \\          
\cline{2-5}
          & \multirow{2}{*}{NAG}  & \multirow{2}{*}{$ \frac{1-\gamma^{m+2-\tau}}{1-\gamma}$} & \multirow{2}{*}{$\mathbf{0}$} &\multirow{2}{*}{No; $\gamma$ is the momentum weight}\\
         &  &                &     \\   
\hline
\multirow{2}{*}{Scaffold} &\multirow{2}{*}{SGD}       & \multirow{2}{*}{ $1$ }    & $\eta m \sum_{r=2}^{t}\mathbf{c}^{(r)} +$  & No; $\mathbf{c}^{(r)}$ is the server control \\   
 &         &     & $\sum_{r=1}^{t-1}\Delta  \mathbf{b}_{j}^{(r)}$ &  variate in global round $r$\\    
\hline
FedProx  &SGD          & $ (1-\lambda\eta)^{m-\tau}$      
 & $ \mathbf{0}$ & $\frac{\lambda}{2}\left\Vert \theta_{k}^{(t,\tau)} - \theta^{(t)}\right\Vert^{2}$\\    
\hline
\multirow{2}{*}{FedDyn}  & \multirow{2}{*}{SGD}   & \multirow{2}{*}{$(1-\lambda\eta)^{m-\tau}$}  & \multirow{2}{*}{$ \left(1 - (1-\lambda\eta)^{m}\right)\sum_{r = 1}^{t-1}\Delta \mathbf{b}_{j}^{(r)}$} & $ \frac{\lambda}{2}\left\Vert \theta_{k}^{(t,\tau)} - \theta^{(t)}\right\Vert^{2}-$\\
 &         &     &   &$ \left\langle \nabla \mathcal{L}_{\text{ce}}^{(t-1,m)}, \theta_{k}^{(t,\tau)} \right\rangle$\\   
 \hline
\multirow{2}{*}{FedDC}  & \multirow{2}{*}{SGD}   & \multirow{2}{*}{$(1-\lambda\eta)^{m-\tau}$}   & $ \left(1 - (1-\lambda\eta)^{m}\right)\sum_{r = 1}^{t-1}\Delta \mathbf{b}_{j}^{(r)} + $ &$\frac{\lambda}{2}\left\Vert \theta_{k}^{(t,\tau)} - ( \theta^{(t)} - \mathbf{d}_{k}^{(t)})\right\Vert^{2} +$\\
 &         &   & $ \frac{1 - (1-\lambda\eta)^{m} }{\lambda\eta m}\left(  \Delta\mathbf{b}_{j}^{(t-1)} -  \Delta\mathbf{B}_{j}^{(t-1)}\right)$ & $ \frac{1}{\eta m}\left\langle \theta_{k}^{(t,\tau)}, \Delta\theta_{k}^{(t-1)} -  \Delta\theta^{(t-1)} \right\rangle$ \\   
\toprule[1pt]
\end{tabular}
\vspace{-0.1 in}
\end{table*}

The prior works on GI attacks in FL focused on FedAvg \citep{fedavg}, where local training pursues minimization of the CE loss, $\mathcal{L}_{\text{ce}}$. Deterioration of the performance of FedAvg observed in non-i.i.d. settings motivated a number of studies \citep{fedprox,scaffold,feddyn,feddc, fedhkd, feddpms, fedmpq} that address the challenge of data heterogeneity by introducing various regularization terms to the local objective function. In particular, those methods consider objectives that are combination of the empirical risk and a regularizer, i.e.,
\begin{equation}
    \mathcal{L}_{\text{local}} = \mathcal{L}_{\text{ce}} + \mathcal{L}_{\text{regularizer}}.
\end{equation}
For such objectives, the local updates of the bias collected by a server are not proportional to the gradients of $\mathcal{L}_{\text{ce}}$, adversely affecting the efficacy of the existing methods that attempt to recover labels from gradients. Furthermore, the existing label recovery methods assume that the local models are updated using SGD optimizers. When optimizers other than SGD are used, local updates are generally not proportional to the gradients of $\mathcal{L}_{\text{ce}}$. To explore such settings, we analyze the expectation of local updates in several milestone non-i.i.d. FL schemes and consider two well-known variants of SGD \citep{sgd}, SGD with momentum (SGDm) and Nesterov accelerated gradient method (NAG). To accommodate these more general cases, we rephrase the expectation of the $j$-th component of the local update of bias $\mathbb{E}\left[\Delta\mathbf{b}_{j}^{(t)}\right]$ in global round $t$ as
\begin{equation}
    \frac{\eta}{|\mathcal{B}|}\sum_{\tau = 1}^{m}\rho^{(\tau)}\left(N_{j}^{(t,\tau)}\sum_{n\not= j} \mathcal{S}_{j,n}^{(t,\tau)} - \sum_{n \not= j} N_{n}^{(t,\tau)}\mathcal{S}_{n,j}^{(t,\tau)}\right) - \mathbf{h}_{j}^{(t)},
\end{equation}
where $\rho^{(\tau)}$ is a constant that depends on hyper-parameters used by different methods and remains constant across global rounds, and $\mathbf{h}_{j}^{(t)}$ is a term capturing historical training information (past local updates). We summarize the values of $\rho^{(\tau)}$ and $\mathbf{h}_{j}^{(t)}$ for different FL schemes in Table~\ref{table1} (derivations provided in Appendix \ref{SGDm_appendix}-\ref{feddc_section}). To the best of our knowledge, this is the first work that studies label recovery attacks in FL schemes beyond FedAvg.
In FL schemes that utilize only the gradient information computed in the current global round (e.g., FedAvg and FedProx) 
$\mathbf{h}_{j}^{t} = \mathbf{0}$, while for the schemes that also
rely on the past gradient information (Scaffold, FedDyn and FedDC)
$\mathbf{h}_{j}^{t} \not = \mathbf{0}$. In any case, $\rho^{(\tau)}$ and $\mathbf{h}_{j}^{(t)}$ are known to the server and may potentially be used for label recovery attacks. In our work, we rely on the procedure for label recovery from local updates discussed in the previous section to run RLU attacks on different FL schemes where $\rho^{(\tau)}$ and $\mathbf{h}_{j}^{t}$ vary from one scheme to another according to Table~\ref{table1}.

\vspace{-0.1 in}
\section{Experiments}
\label{experiments}
\subsection{Setups}
\label{setup}
We evaluate the performance of RLU on a classification task using a variety of model architectures including LeNet-5 \citep{lenet}, VGG-16 \citep{vgg} and ResNet-50 \citep{resnet}, and four benchmark datasets including SVHN \citep{svhn}, CIFAR10, CIFAR100 and Tiny-ImageNet \citep{tiny}. Throughout these experiments we employ a number of activation functions including ReLU, Tanh, ELU \citep{elu}, SELU \citep{selu} and SiLU \citep{silu} to further  evaluate robustness of our proposed method. To simulate diverse FL scenarios, we follow the strategy in \citep{bayesian} and utilize Dirichlet distribution with a concentration parameter $\alpha$, controlling the level of data heterogeneity across $10$ data partitions owned by 10 clients. Unless specified otherwise, the models are trained using SGD optimizers. The auxiliary dataset $\mathcal{A}$ contains $100$ samples per class. Further experimental details are provided in Appendix \ref{settings_appendix}.
\begin{table*}[t]
\caption{Comparison of class-level accuracy (cAcc) and instance-level accuracy (iAcc) of various methods in label recovery attacks on \textbf{untrained} models. The concentration parameter $\alpha$ controlling data partitioning is set to $0.5$ in experiments on SVHN and CIFAR10, and to $0.1$ on CIFAR100 and Tiny-ImageNet. The optimizer used in all experiments is SGD.} 
\centering
\small
\begin{tabular}{cccccccccccc}
\bottomrule[1pt]
\label{table2}
\multirow{2}{*}{Model }   & \multirow{2}{*}{Dataset} & Batch & \multirow{2}{*}{Activation}   & 
 \multicolumn{2}{c}{LLG+} & \multicolumn{2}{c}{ZLG+} & \multicolumn{2}{c}{iRLG} & \multicolumn{2}{c}{\textbf{RLU (ours)}}\\
\cline{5-12}
   &   & Size    & & cAcc & iAcc & cAcc & iAcc  & cAcc & iAcc & cAcc & iAcc \\
\hline
   &   &       \multicolumn{8}{c}{single local epoch $m$ = 1}  &  &     \\

\hline   
\multirow{2}{*}{LeNet-5 } & \multirow{2}{*}{SVHN}     & \multirow{2}{*}{32} &ReLU&  0.980  & \cellcolor[HTML]{EFEFEF} 0.984 &0.810  & \cellcolor[HTML]{EFEFEF} 0.906   & 0.998 & \cellcolor[HTML]{EFEFEF} \textbf{1.000}&  \textbf{1.000}   & \cellcolor[HTML]{EFEFEF} \textbf{1.000}          \\   
&   &  &Tanh&0.199 & \cellcolor[HTML]{EFEFEF} 0.063 &0.314  &\cellcolor[HTML]{EFEFEF} 0.831 &0.994& \cellcolor[HTML]{EFEFEF} \textbf{1.000}&  \textbf{1.000}   & \cellcolor[HTML]{EFEFEF} \textbf{1.000}  \\
\hline
\multirow{4}{*}{Vgg-16 } & \multirow{2}{*}{CIFAR10}     & \multirow{2}{*}{64} &ReLU&0.982  & \cellcolor[HTML]{EFEFEF} 0.981  &0.907 &\cellcolor[HTML]{EFEFEF} 0.982 &0.961 & \cellcolor[HTML]{EFEFEF} 0.979 &\textbf{1.000 }   &\cellcolor[HTML]{EFEFEF} \textbf{1.000 }           \\   
&   &  &ELU&0.845 & \cellcolor[HTML]{EFEFEF} 0.925 & 0.962 & \cellcolor[HTML]{EFEFEF} 0.932 &0.920 &\cellcolor[HTML]{EFEFEF} 0.936 & \textbf{0.994} & \cellcolor[HTML]{EFEFEF} \textbf{0.996}  \\
\cline{2-12}
& \multirow{2}{*}{CIFAR100}     &\multirow{2}{*}{256} &ReLU&   0.960   & \cellcolor[HTML]{EFEFEF} 0.932&0.888  &\cellcolor[HTML]{EFEFEF} 0.943  & 0.981 & \cellcolor[HTML]{EFEFEF} 0.992  &\textbf{1.000 }   & \cellcolor[HTML]{EFEFEF} \textbf{1.000 }    \\   
&   &  &SELU &0.267  &\cellcolor[HTML]{EFEFEF} 0.104   &0.893     & \cellcolor[HTML]{EFEFEF} 0.855      &0.958     & \cellcolor[HTML]{EFEFEF} 0.982  &\textbf{1.000 } & \cellcolor[HTML]{EFEFEF} \textbf{1.000 }   \\
\hline
\multirow{2}{*}{ResNet-50} & \multirow{2}{*}{Tiny}     & \multirow{2}{*}{256} &ReLU&0.998   & \cellcolor[HTML]{EFEFEF} 0.938    &0.880 & \cellcolor[HTML]{EFEFEF} 0.666   &0.992 & \cellcolor[HTML]{EFEFEF} 0.995 &\textbf{1.000}   & \cellcolor[HTML]{EFEFEF} \textbf{1.000}    \\   
&   &  &SiLU&0.953   & \cellcolor[HTML]{EFEFEF} 0.794    &0.847   & \cellcolor[HTML]{EFEFEF} 0.534  &0.989 & \cellcolor[HTML]{EFEFEF} \textbf{1.000} &\textbf{1.000} & \cellcolor[HTML]{EFEFEF} \textbf{1.000}\\
\hline
   &   &       \multicolumn{8}{c}{multiple local epochs $m = 10$}  &  &     \\
\hline
\multirow{2}{*}{LeNet-5 } & \multirow{2}{*}{SVHN}     & \multirow{2}{*}{32} &ReLU& 0.863  &\cellcolor[HTML]{EFEFEF} 0.934   &0.496 & \cellcolor[HTML]{EFEFEF} 0.644 &0.996 & \cellcolor[HTML]{EFEFEF} 0.997  & \textbf{1.000}  & \cellcolor[HTML]{EFEFEF} \textbf{1.000}        \\   
&   &  &Tanh&0.218 & \cellcolor[HTML]{EFEFEF} 0.164 &0.551  & \cellcolor[HTML]{EFEFEF} 0.699 &0.976 & \cellcolor[HTML]{EFEFEF}0.996 & \textbf{1.000}   & \cellcolor[HTML]{EFEFEF} \textbf{1.000}  \\
\hline
\multirow{4}{*}{Vgg-16 } & \multirow{2}{*}{CIFAR10}     & \multirow{2}{*}{64} &ReLU&0.674  & \cellcolor[HTML]{EFEFEF} 0.875   & 0.671 & \cellcolor[HTML]{EFEFEF} 0.866 & 0.843 & \cellcolor[HTML]{EFEFEF} 0.717  &\textbf{0.845}  & \cellcolor[HTML]{EFEFEF} \textbf{0.946}   \\   
&   &  &ELU&0.688 & \cellcolor[HTML]{EFEFEF} 0.812 &0.682 & \cellcolor[HTML]{EFEFEF} 0.798 &0.808   & \cellcolor[HTML]{EFEFEF} 0.650 &\textbf{0.866}     &\cellcolor[HTML]{EFEFEF} \textbf{0.923}  \\
\cline{2-12}
& \multirow{2}{*}{CIFAR100}     &\multirow{2}{*}{256} &ReLU     &0.902  &\cellcolor[HTML]{EFEFEF} 0.948&0.693 &\cellcolor[HTML]{EFEFEF}0.929  &0.463  &\cellcolor[HTML]{EFEFEF} 0.902  & \textbf{0.922}  & \cellcolor[HTML]{EFEFEF} \textbf{0.981}  \\   
&   &  &SELU &0.072  &\cellcolor[HTML]{EFEFEF} 0.014   &0.400     & \cellcolor[HTML]{EFEFEF} 0.840   &0.466  &\cellcolor[HTML]{EFEFEF} 0.689 &\textbf{0.843} & \cellcolor[HTML]{EFEFEF} \textbf{0.904} \\
\hline
\multirow{2}{*}{ResNet-50} & \multirow{2}{*}{Tiny}     & \multirow{2}{*}{256} &ReLU&0.929   &\cellcolor[HTML]{EFEFEF} 0.921    &0.643 & \cellcolor[HTML]{EFEFEF} 0.723   &0.792 &\cellcolor[HTML]{EFEFEF} 0.937 & \textbf{0.957}  &\cellcolor[HTML]{EFEFEF} \textbf{0.976}     \\   
&   &  &SiLU&0.922   &\cellcolor[HTML]{EFEFEF} 0.840    &0.589   &\cellcolor[HTML]{EFEFEF} 0.759  &0.983 &\cellcolor[HTML]{EFEFEF} 0.984  &\textbf{0.991} & \cellcolor[HTML]{EFEFEF} \textbf{0.996}\\
\toprule[1pt]
\end{tabular}
\end{table*}

\vspace{-0.1 in}
\subsection{Baselines and Evaluation Metrics}
\label{section_metrics}
We compare our proposed RLU to three state-of-the-art methods: LLG \citep{llg}, ZLG \citep{zlg} and iRLG \citep{irlg}, all capable of recovering repeated labels in a batch. For fairness, we compare RLU to LLG+ and ZLG+; the latter two utilize the same auxiliary dataset $\mathcal{A}$ as RLU to achieve improved performance. Following the strategy of iRLG, we quantify the performance of a label recovery attack using two metrics: (1) \emph{class-level accuracy (cAcc)}: the proportion of correctly recovered classes; (2) \emph{instance-level accuracy (iAcc)}: the proportion of correctly recovered labels. Details of computing cAcc and iAcc are provided in Appendix~\ref{metrics}. We recover batch labels from the local updates computed on clients' local dataset and report the average cAcc and iAcc.

\vspace{-0.1 in}
\subsection{Attack on Untrained Models}

\label{untrained}
A randomly initialized untrained model may be extremely vulnerable to label recovery attacks; for instance, if the server knows how the training is initialized, it does not even need an auxiliary data set to infer the parameters (e.g., $\boldsymbol{\mu}_{n}$) used in the attack. To compare the performance of baseline methods with that of RLU, we conduct comprehensive experiments on untrained models across various architectures, datasets, batch-sizes and activation functions. The results, reported in Table~\ref{table2}, demonstrate that RLU outperforms the baselines in all settings. When local training consists of a single epoch, RLU achieves near-perfect accuracy across the board in terms of both cAcc and iAcc; iRLG is a close second, outperforming other baselines. Since LLG+ assumes non-negative activation functions, its performance deteriorates significantly with Tanh and SELU. On Tiny-ImageNet, ZLG+ performs the worst among the four methods, achieving under $70\%$ iAcc.

When the clients run $m = 10$ local epochs, performance of all methods deteriorates (as expected based on the discussion in Section \ref{multiple_epochs}). Nevertheless, the results in Table~\ref{table2} show that RLU still outperforms the baselines, maintaining at least $84\%$ cACC and $90\%$ iAcc on all datasets, architectures and activation functions. While iRLG maintains solid performance on SVHN and Tiny-ImageNet, it performs significantly worse on CIFAR10 and CIFAR100 (iAcc falls below $70\%$, cAcc drops below $50\%$). The results of LLG+ follow the same pattern exhibited in single epoch settings, while ZLG+ experiences significant performance deterioration on SVHN. For consistency and a comparison with the results in Table~\ref{table2}, unless stated otherwise, in the remainder of this section the number of local epochs $m$ is set to $10$.

\vspace{- 0.1 in}
\subsection{Attack on Trained Models}
\label{trained}
\begin{figure*}[t] 
    \centering
	  \subfloat[SVHN]{
       \includegraphics[width=0.33\linewidth]{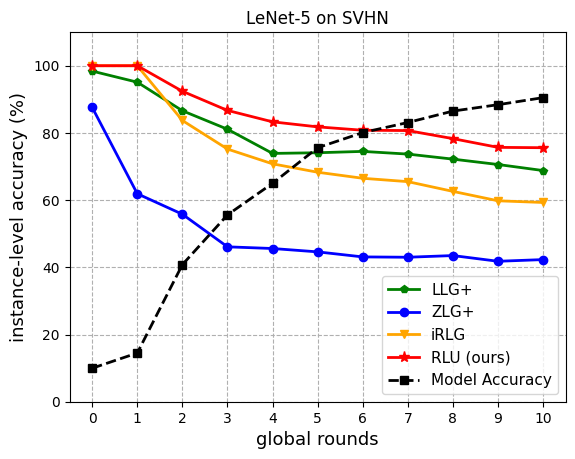}}
	  \subfloat[CIFAR10]{
        \includegraphics[width=0.33\linewidth]{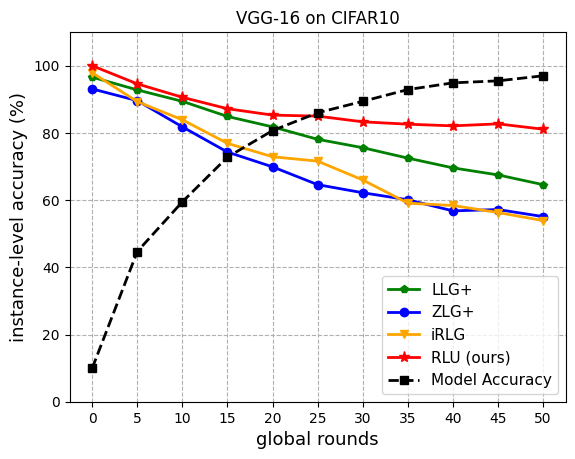}}
	  \subfloat[CIFAR100]{
        \includegraphics[width=0.33\linewidth]{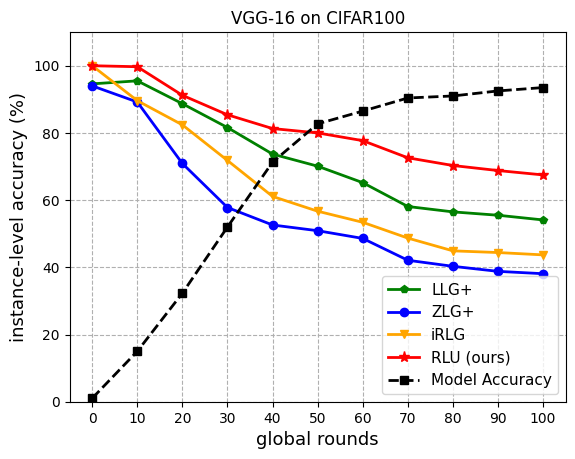}}
	\caption{Instance-level accuracy of different attack methods deteriorates as training progresses. Each point on the black dashed curve indicates the training accuracy of the global model in each global round.}
\label{trained_models} 
\end{figure*}
Since the accuracy of the global model improves as the training proceeds, comparing methods in terms of attacks on untrained model becomes no longer meaningful. As discussed in Section \ref{charateristic}, distribution of output logits depends on the accuracy of the model which is parameterized by mean $\boldsymbol{\mu}_{n}$ and covariance $\boldsymbol{\Sigma}_{n}$. When the model becomes highly accurate, magnitudes of local gradients (updates) start vanishing; this, in turn, causes high sensitivity to noise and a large error in estimating the labels. All of the methods in Table~\ref{table2} experience performance deterioration as the model accuracy increases. To compare their robustness, we conduct extensive experiments and evaluate performance of different attack methods at different stages of training. As shown in Fig.~\ref{trained_models}, RLU outperforms the baselines at all training stages, achieving $80\%$ iAcc when the model accuracy reaches $80\%$. For the global model achieving $90\%$ training accuracy, iAcc of iRLG reaches $75.6 \%$, $83.3\%$ and $72.6\%$ on SVHN, CIFAR10 and CIFAR100, respectively. The iAcc of ZLG+ is $33.3\%$, $21.1\%$ and $30.1\%$ lower than RLU's, while iRLG's trails RLU by $16.3 \%$, $17.3 \%$ and $23.9 \%$ on these three datasets, respectively. The iAcc performance of LLG+ is better than that of other prior methods but still falls significantly behind  RLU's. 

\subsection{Effect of Data Heterogeneity}
\label{heterogeneity}
\begin{table}[t]
\caption{Experiments on CIFAR10 comparing iAcc of different methods on data partitions generated by varying $\alpha$.} 
\centering
\begin{tabular}{cccccc}
\bottomrule[1pt]
\label{table3}
    & \multicolumn{5}{c}{Data heterogeneity $\alpha$}\\
\hline  
Schemes  & 0.05 & 0.1 & 0.5 &  1 & 5   \\
\hline   
LLG+  & 0.766  & 0.831  & 0.882  & 0.905 &  0.941  \\   
ZLG+  &0.747   & 0.812  & 0.867  & 0.897  & \textbf{0.944}   \\  
iRLG  &0.449   &0.630  & 0.715  & 0.766 & 0.820 \\  
RLU  &\textbf{0.961}    &\textbf{0.947}  &\textbf{0.944}  & \textbf{0.943} &0.931  \\  
\toprule[1pt]
\end{tabular}
\vspace{-0.1 in}
\end{table}
\begin{table*}[t]
\caption{Instance-level accuracy achieved by various methods in label recovery attacks on different FL schemes.} 
\centering
\small
\begin{tabular}{cc|cccc|cccc}
\bottomrule[1pt]
\label{table4}
Schemes    & Dataset &
LLG+ & ZLG+& iRLG & \textbf{RLU (ours)} &
LLG+ & ZLG+& iRLG & \textbf{RLU (ours)}\\
\hline
Hyper-parameter:   &    & \multicolumn{4}{c|}{$\lambda = 0.5$} & \multicolumn{4}{c}{$\lambda = 5$} \\
\hline   
\multirow{3}{*}{FedProx } & SVHN    & 0.917& 0.656& 0.982& \textbf{0.991} &0.079 &0.566 &0.881   &\textbf{0.973}   \\   
 & CIFAR10      &0.747&0.795  &0.691  &\textbf{0.938} &0.665 & 0.682& 0.616 &\textbf{0.899}      \\   
  & CIFAR100      &0.805   &0.791   &0.594   & \textbf{0.930} & 0.659 &0.626 &0.493 & \textbf{0.929}      \\   
\hline
Hyper-parameter:   &    & \multicolumn{4}{c|}{$\gamma = 0.1$} & \multicolumn{4}{c}{$\gamma = 0.9$} \\
\hline
\multirow{3}{*}{SGDm } & SVHN      & 0.951& 0.520  &0.819 & \textbf{0.987}    &0.737   &0.870  &0.114  &\textbf{0.916} \\
 & CIFAR10  &0.884 &0.872  &0.919   &\textbf{0.935}    &0.883  &0.871 &0.450 & \textbf{0.928} \\
  & CIFAR100  &0.855 &0.826  &0.856   &\textbf{0.915}    &0.864  &0.859 &0.755 &\textbf{0.882} \\
\toprule[1pt]
\end{tabular}
\end{table*}
Data heterogeneity is one the main challenges in real-world applications of federated learning. Prior works on label reconstruction evaluated their proposed methods on i.i.d. data, but the evaluation on non-i.i.d. data has remained largely unexplored. To this end, we benchmark the methods considered in this paper on CIFAR10 data partitions generated for varied values of $\alpha = \{0.05, 0.1, 0.5, 1, 5\}$ (smaller $\alpha$ corresponds to higher level of heterogeneity). We visualize the generated data partitions in Appendix \ref{visualization}. Table~\ref{table3} shows that iAcc of the three baselines monotonically decreases with the level of data heterogeneity. On the other hand, RLU demonstrate a great degree of robustness as it maintains high iAcc across the board; in particular, RLU achieves $93\%$ or higher instance-level accuracy in all settings, including at the highest level of data heterogeneity ($\alpha = 0.05$).
\begin{figure}[t] 
    \centering
    \includegraphics[width= 0.9 \linewidth]{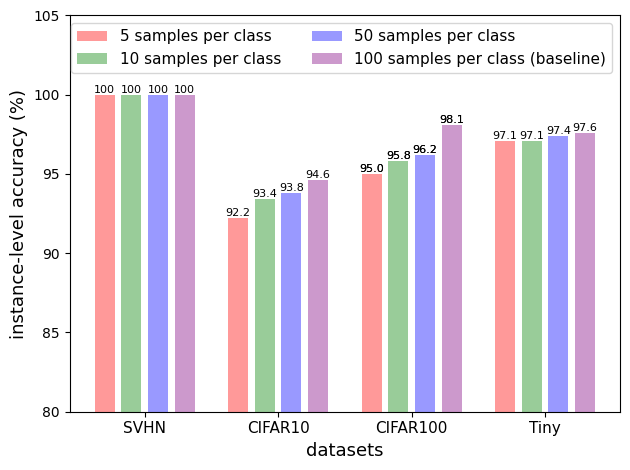}
	\caption{The iAcc of RLU utilizing auxiliary dataset $\mathcal{A}$ as the number of samples per class varies.}
\label{size_results} 
\vspace{-0.1 in}
\end{figure}

\subsection{Effect of the Size of Auxiliary Dataset $\mathcal{A}$}
\label{auxiliary}

\begin{figure*}[t] 
    \centering
	  \subfloat[Ground Truth]{
       \includegraphics[width=0.23\linewidth]{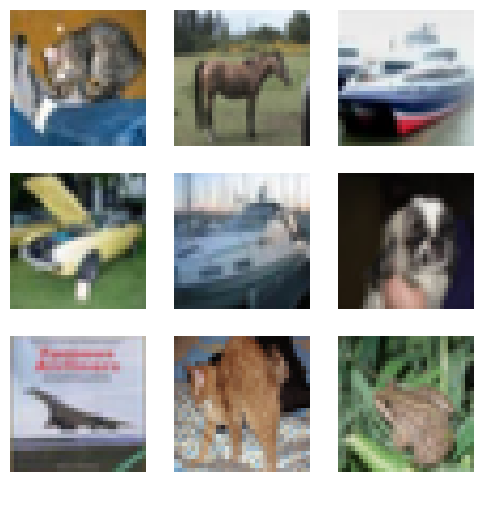}}
       \hspace{0.3in}
        \subfloat[IG (Joint Optimization)]{
        \includegraphics[width=0.23\linewidth]{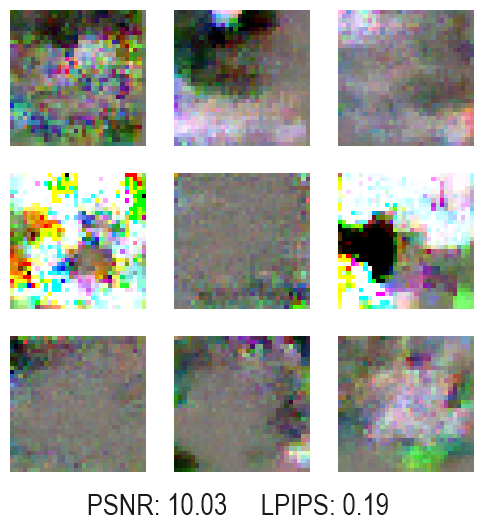}}
       \hspace{0.3in}
	  \subfloat[Improved by RLU]{
        \includegraphics[width=0.23\linewidth]{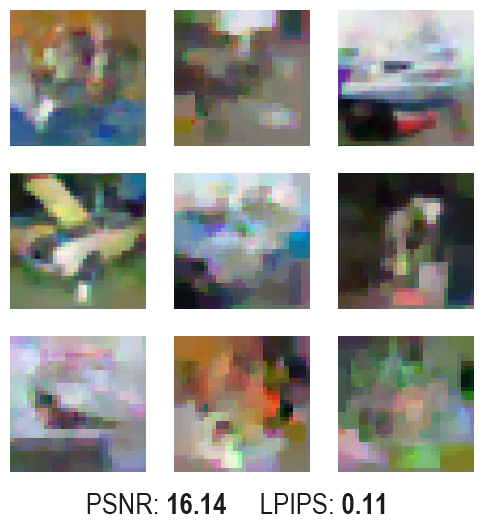}}
	\caption{Batch image reconstruction (batch size set to 9) on CIFAR10 compared to IG \citep{geiping}. We select the best reconstructed batch for visualization and display the average metrics of the selected batches.}
\label{IG_visualization} 
\end{figure*}

As previously discussed, RLU needs an auxiliary dataset $\mathcal{A}$ to estimate moments of the output logits distribution. In the benchmarking experiments presented thus far, we used $100$ samples for each class in $\mathcal{A}$. To analyze the effect of the size of auxiliary dataset on the performance of RLU, we conduct $4$ sets of experiments that utilize $4$ auxiliary datasets with: (1) $5$ samples per class; (2) $10$ samples per class; (3) $50$ samples per class; and (4) $100$ samples per class. As shown in Fig.~\ref{size_results}, there appears to be no significant performance degradation due to reduction of the auxiliary data set size. When using the smallest among the auxiliary sets, on SVHN and Tiny the proposed RLU achieves performance close to the baseline. The largest performance gap is on CIFAR10 and CIFAR100, and even there the gap is only $2.4\%$ and $3.1\%$, respectively. Therefore, RLU exhibits robustness with respect to the variations in the size of the auxiliary data set used by the server. 

\subsection{Attacks on Different FL Schemes}
\label{different_fl}
To the best of our knowledge, prior works evaluate their methods only on FedAvg. While FedAvg is indeed the oldest and perhaps the most widely used FL scheme, a number of other FL schemes has grown to prominence yet remains largely unexplored in the context of label recovery. As discussed in Section \ref{beyond_fedavg}, we provide a framework to conduct label recovery attacks on several regularization-based FL schemes that may be using various optimizers. Table~\ref{table4} shows the superior performance of RLU in the experiments on FedProx and SGDm. As can be seen from the table, in the experiments where $\lambda = 0.5$ and $\gamma = 0.1$ (which leads to $\rho^{(\tau)} \approx 1$), the three baselines achieve performance similar to that in their attacks on FedAvg. However, as $\rho^{(\tau)}$ deviates from $1$ when $\lambda = 5$ and $\gamma = 0.9$, performance of the baselines severely deteriorates while RLU maintain its iAcc of at least $88\%$. 

\subsection{Improved Gradient Inversion Attacks with RLU}
The outstanding performance of RLU on label recovery can improve the gradient inversion attacks in federated learning. In the prior work IG \citep{geiping}, an HBC server performs joint optimization of the reconstructed labels $\mathbf{y}^{\prime}$ and images $\mathbf{x}^{\prime}$, which typically results in slow convergence and poor quality of the reconstructed images. We conduct gradients inversion attack experiments on CIFAR10, where we use RLU to estimate labels $\mathbf{y}^{\prime}$ from local updates and only optimize the reconstructed images $\mathbf{x}^{\prime}$ according to Eq. \ref{ig}. To quantitatively characterize the quality of reconstructed images, we compute the peak signal-to-noise ratio (PSNR) and the learned perceptual image patch similarity (LPIPS) \citep{LPIPS} of the reconstructed images in each batch. As illustrated in Fig. \ref{IG_visualization}, images reconstructed with the help of RLU have higher PSNR and lower LPIPS, indicating smaller distance to the original images. 
\label{different}

\section{Conclusion}
\label{conclusion}
In this paper, we studied label recovery attacks on federated learning systems in which clients send their local model updates to the server for aggregation. We developed RLU, a novel label recovery method which solves a least-square problem constructed by examining the correlation between the number of samples of each label type and local updates of the output layer. We extended the proposed framework to real-world scenarios involving well-trained models, multiple local epochs, high levels of data heterogeneity and various local objective functions, and provided theoretical analysis of RLU in different FL schemes. Comprehensive experiments on four datasets, three model architectures and six activation functions demonstrate consistently high accuracy, robustness and universality of RLU. Moreover, gradients inversion attack experiments illustrate that utilizing RLU may significantly improve quality of the reconstructed data in term of two widely-used metrics, PSNR and LPIPS. Future work will include exploring defense mechanism that may help ameliorate safety concerns caused by RLU.


\bibliography{icml2024}
\bibliographystyle{icml2024}

\newpage
\appendix
\onecolumn
\section{Proof of Analytical Results}
\subsection{Derivation of the gradient of the output layer's bias $\mathbf{b}$}
\label{gradient_bias}
Given a batch of samples $(\mathbf{x}^{(i)}, y^{(i)}) \in \mathcal{B}$, the cross-entrop loss can be computed as
\begin{equation}
\label{celoss_appendix}
    \mathcal{L}_{\text{ce}} = = \frac{1}{|\mathcal{B}|}\sum_{i = 1}^{|\mathcal{B}|}\mathcal{L}_{\text{ce}}^{(i)} = -\frac{1}{|\mathcal{B}|}\sum_{i = 1}^{|\mathcal{B}|}\log\frac{\exp(\mathbf{q}_{y^{(i)}}^{(i)})}{\sum_{n = 1}^{N} \exp(\mathbf{q}_{n}^{(i)})}, 
\end{equation}

\begin{equation}
\label{fc}
    \mathbf{q}_{j}^{(i)} = \sum_{l=1}^{L} \mathbf{W}_{j,l}\mathbf{z}_{l}^{(i)} + \mathbf{b}_{j},
\end{equation}
where $N$ is the number of classes; $\mathbf{q}_{j}^{(i)}$ denotes the $j$-th component of output logits given sample $\mathbf{x}^{(i)}$; $\mathbf{W}_{j,l}$ is the $(j,l)$-element of weights of the output layer; $\mathbf{b}_{j}$ is the $j$-th component of bias of the output layer; $\mathbf{z}^{(i)}$ is the embedding of data $\mathbf{x}^{(i)}$; $L$ is the dimension of embedding space. The gradient of the bias $\mathbf{b}_{j}$ given sample $(\mathbf{x}^{(i)}, y^{(i)})$ can be computed using the chain rule as
\begin{equation}
\label{chain}
\frac{\partial \mathcal{L}_{\text{ce}}^{(i)}}{\partial \mathbf{b}_{j}} = \frac{\partial \mathcal{L}_{\text{ce}}^{(i)}}{\partial \mathbf{Q}} \cdot \frac{\partial \mathbf{Q}}{\partial \mathbf{q}_{j}^{(i)}} \cdot \frac{\partial \mathbf{q}_{j}^{(i)}}{\partial \mathbf{b}_{j}} ,
\end{equation}
where
\begin{equation}
    \mathbf{Q} = \frac{\exp(\mathbf{q}_{y^{(i)}}^{(i)})}{\sum_{n = 1}^{N} \exp(\mathbf{q}_{n}^{(i)})}.
\end{equation}
To start, note that 
\begin{equation}
    \frac{\partial \mathcal{L}_{\text{ce}}^{(i)}}{\partial \mathbf{Q}} = -\frac{1}{\mathbf{Q}}, \frac{\partial \mathbf{q}_{j}^{(i)}}{\partial \mathbf{b}_{j}} = 1.
\end{equation}
If $j = y^{(i)}$, we obtain
\begin{equation}
\label{grad_true}
     \frac{\partial \mathbf{Q}}{\partial \mathbf{q}_{j}^{(i)}} = \frac{\exp\left(\mathbf{q}^{(i)}_{y^{(i)}}\right)\left(\sum_{n = 1}^{N}\exp\left(\mathbf{q}^{(i)}_{n}\right)\right) - \exp\left(\mathbf{q}^{(i)}_{y^{(i)}}\right)^{2}}{\left(\sum_{n = 1}^{N}\exp\left(\mathbf{q}^{(i)}_{n}\right)\right)^{2}} = \frac{\sum_{n \not = y^{(i)}}\exp\left(\mathbf{q}^{(i)}_{n}\right)}{\sum_{n = 1}^{N}\exp\left(\mathbf{q}^{(i)}_{n}\right)} \cdot \mathbf{Q},
\end{equation}
while if $j \not = y^{(i)}$,
\begin{equation}
\label{grad_false}
     \frac{\partial \mathbf{Q}}{\partial \mathbf{q}_{j}^{(i)}} = - \frac{  \exp\left(\mathbf{q}^{(i)}_{y^{(i)}}\right) \exp\left(\mathbf{q}^{(i)}_{i}\right)}{\left(\sum_{n = 1}^{N}\exp\left(\mathbf{q}^{(i)}_{n}\right)\right)^{2}} = -\frac{\exp\left(\mathbf{q}^{(i)}_{i}\right)}{\sum_{n = 1}^{N}\exp\left(\mathbf{q}^{(i)}_{n}\right)} \cdot \mathbf{Q}.
\end{equation}
By substituting Eq.~\ref{grad_true} and Eq.~\ref{grad_false} in Eq.~\ref{chain}, we obtain
\begin{equation}
\label{gradient_bias_appendix}
    \nabla\mathbf{b}_{j}^{(i)} = \left\{
    \begin{aligned}
        &\frac{\exp(\mathbf{q}_{j}^{(i)})}{\sum_{n = 1}^{N} \exp(\mathbf{q}_{n}^{(i)})}, \text{ if } j \not = y^{(i)}, \\
        & -\frac{\sum_{n \not = j}\exp(\mathbf{q}_{n}^{(i)})}{\sum_{n = 1}^{N} \exp(\mathbf{q}_{n}^{(i)})}, \text{ if } j  = y^{(i)}.\\
    \end{aligned}
    \right.
\end{equation}
$\hfill \blacksquare$

\subsection{Derivation of the gradient of the output layer's weight $\mathbf{W}$}
\label{gradient_weight}
According to Eq. \ref{celoss_appendix} and Eq. \ref{fc}, gradients of the weight $\mathbf{W}_{j,l}$ given sample $(\mathbf{x}^{(i)}, \mathbf{y}^{(i)})$ can be computed as
\begin{equation}
\label{chain_w}
\frac{\partial \mathcal{L}_{\text{ce}}^{(i)}}{\partial \mathbf{W}_{j,l}} = \frac{\partial \mathcal{L}_{\text{ce}}^{(i)}}{\partial \mathbf{Q}} \cdot \frac{\partial \mathbf{Q}}{\partial \mathbf{q}_{j}^{(i)}} \cdot \frac{\partial \mathbf{q}_{j}^{(i)}}{\partial \mathbf{W}_{j,l}}.
\end{equation}
Since, as discussed in Section \ref{gradient_bias}, we know $\frac{\partial \mathcal{L}_{\text{ce}}^{(i)}}{\partial \mathbf{Q}}$ and $\frac{\partial \mathbf{Q}}{\partial \mathbf{q}_{j}^{(i)}}$, we have $\frac{\partial \mathbf{q}_{j}^{(i)}}{\partial \mathbf{W}_{j,l}} = \mathbf{z}_{j}^{(i)}$.
Finally, we obtain
\begin{equation}
\label{gradient_weight_appendix}
    \nabla\mathbf{W}_{j,l}^{(i)} = \nabla\mathbf{b}_{j}^{(i)} \cdot \mathbf{z}_{l}^{(i)}.
\end{equation}

$\hfill \blacksquare$

\subsection{Derivation of the expected local update of the output layer's bias $\mathbf{b}$ (single epoch)}
\label{expectation_bias_appendix}
Suppose mini-batch stochastic gradient descent (SGD) is used as the optimizer in FL training; the local update computed in a single epoch can be found as
\begin{equation}
    \Delta \mathbf{b}_{j} = -\eta\nabla\mathbf{b}_{j} = -\frac{\eta}{|\mathcal{B}|}\sum_{i=1}^{|\mathcal{B}|}\nabla\mathbf{b}_{j}^{(i)}.
\end{equation}
Taking expectation of both sides yields
\begin{equation}
    \begin{aligned}
\mathbb{E}\left[\Delta \mathbf{b}_{j}\right] &= -\frac{\eta}{|\mathcal{B}|}\sum_{i=1}^{|\mathcal{B}|} \mathbb{E}\left[\nabla\mathbf{b}_{j}^{(i)}\right]\\
&=  \frac{\eta}{|\mathcal{B}|}\sum_{i=1}^{|\mathcal{B}|}\left(\mathbb{I}\{j = y^{(i)}\}\mathbb{E}\left[\frac{\sum_{n \not = j}\exp({\mathbf{q}^{(i)}_{n}})}{\sum_{n = 1}^{N}\exp({\mathbf{q}^{(i)}_{n}})}\right]  -  \mathbb{I}\{j \not = y^{(i)}\} \mathbb{E}\left[ \frac{\exp({\mathbf{q}^{(i)}_{j}})}{\sum_{n = 1}^{N}\exp({\mathbf{q}^{(i)}_{n}})}\right]\right)\\
&= \frac{\eta}{|\mathcal{B}|} \left( N_{j}\sum_{n \not = j}\mathbb{E}_{(\mathbf{x},y) \sim \mathcal{D}_{k}^{(j)}}\left[\mathbf{s}_{n}(\mathbf{x})\right]  - \sum_{n \not = j} N_{n}\mathbb{E}_{(\mathbf{x},y) \sim \mathcal{D}_{k}^{(n)}} \left[\mathbf{s}_{j}(\mathbf{x})\right] \right)\\
&= \frac{\eta}{|\mathcal{B}|} \left( N_{j}\sum_{n \not = j}\mathcal{S}_{j,n}  -  \sum_{n \not = j} N_{n}\mathcal{S}_{n,j}\right).\\
    \end{aligned}
\end{equation}
Let $\mathcal{D}_{k}^{(-j)}$ denote the subset of local dataset $\mathcal{D}_{k}$ that excludes the samples with label $j$, and let $\mathcal{S}_{j} = \mathbb{E}_{(\mathbf{x},y)\sim \mathcal{D}_{k}^{(-j)}}\left[\mathbf{s}_{j}(\mathbf{x})\right]$ (can be approximated by $\mathcal{S}_{j} \approx \frac{1}{N-1}\sum_{n \not = j} \mathcal{S}_{n,j}$). It follows that
\begin{equation}
\label{sum_Sn}
\begin{aligned}
    \sum_{j=1}^{N}\mathcal{S}_{j} &=  \sum_{j=1}^{N}\mathbb{E}_{(\mathbf{x},y)\sim \mathcal{D}_{k}^{(-j)}}\left[\mathbf{s}_{j}(\mathbf{x})\right]\\
    &\approx \frac{1}{N-1}\sum_{j=1}^{N}\underbrace{\sum_{n\not=j}\frac{1}{|\mathcal{D}_{k}|\cdot P_{n}}\sum_{i=1}^{|\mathcal{D}_{k}|}\mathcal{I}\{y^{(i)}  = n\}\frac{\exp(\mathbf{q}_{j}^{(i)})}{\sum_{c=1}^{N} \exp(\mathbf{q}_{c}^{(i)})}}_{\text{reformulate}}\\
    &=  \frac{1}{N-1}\sum_{j=1}^{N}\frac{1}{|\mathcal{D}_{k}|\cdot P_{j}}\sum_{i=1}^{|\mathcal{D}_{k}|}\mathcal{I}\{y^{(i)}  = j\}\frac{\sum_{n\not=j}\exp(\mathbf{q}_{j}^{(i)})}{\sum_{n=1}^{N} \exp(\mathbf{q}_{n}^{(i)})}\\
    &= \frac{1}{N-1}\sum_{j=1}^{N}\left(1 - \frac{1}{|\mathcal{D}_{k}|\cdot P_{j}}\sum_{i=1}^{|\mathcal{D}_{k}|}\mathcal{I}\{y^{(i)} = j\}\frac{\exp(\mathbf{q}_{j}^{(i)})}{\sum_{n=1}^{N} \exp(\mathbf{q}_{n}^{(i)})}\right)\\
    &= \frac{N}{N-1} - \frac{1}{(N-1)|\mathcal{D}_{k}|}\sum_{j=1}^{N}\sum_{i=1}^{|\mathcal{D}_{k}|}\frac{\exp(\mathbf{q}_{y^{(i)}}^{(i)})}{\sum_{n=1}^{N} \exp(\mathbf{q}_{n}^{(i)})} \cdot \frac{\mathcal{I}\{y^{(i)} = j\}}{P_{j}},
\end{aligned}
\end{equation}
where $P_{j}$ is the proportion of samples with label $j$ in dataset $\mathcal{D}_{k}$. Note that the second term in Eq.~\ref{sum_Sn} is positively correlated to $\mathcal{L}_{\text{ce}}$ specified in Eq.~\ref{celoss_appendix}. Given an untrained neural network model, we have
\begin{equation}
    \frac{\exp(\mathbf{q}_{y^{(i)}}^{(i)})}{\sum_{n=1}^{N} \exp(\mathbf{q}_{n}^{(i)})} \approx \frac{1}{N}, \forall y^{(i)} \in [N],
\end{equation}
and thus we obtain
\begin{equation}
    \sum_{j=1}^{N}\mathcal{S}_{j} \approx \frac{N}{N-1} - \frac{1}{N-1} \sum_{j=1}^{N} \frac{1}{N}  \frac{|\mathcal{D}_{k}|\cdot P_{j}}{|\mathcal{D}_{k}|\cdot P_{j}} = \frac{N-1}{N-1} = 1.
\end{equation}
Assume that after $T$ global rounds $\mathcal{L}_{ce}^{\star} = 0$; it follows that
\begin{equation}
    \frac{\exp(\mathbf{q}_{y^{(i)}}^{(i)})}{\sum_{n=1}^{N} \exp(\mathbf{q}_{n}^{(i)})} = 1, \forall y^{(i)} \in [N],
\end{equation}
and thus we obtain
\begin{equation}
     \sum_{j=1}^{N}\mathcal{S}_{j} \approx \frac{N}{N-1} - \frac{1}{N-1} \sum_{j=1}^{N} 1 \cdot \frac{|\mathcal{D}_{k}|\cdot P_{j}}{|\mathcal{D}_{k}|\cdot P_{j}} = \frac{N}{N-1} -  \frac{N}{N-1}  = 0.
\end{equation}
$\hfill \blacksquare$

\subsection{Derivation of the expected local update using SGD with momentum (SGDm)}
\label{SGDm_appendix}
In each local epoch $\tau$, SGD with momentum \citep{sgd} updates the model parameters $\theta^{(t,\tau)}_{k}$ of client $k$ according to 
\begin{equation}
\label{momentum_appendix}
    \mathbf{v}^{(t,\tau)}_{k} \xleftarrow{} \gamma \mathbf{v}^{(t,\tau-1)}_{k} + \nabla\mathcal{L}_{\text{ce}}^{(t,\tau)},
\end{equation}
\begin{equation}
    \theta^{(t,\tau)}_{k} \xleftarrow{} \theta^{(t,\tau-1)}_{k} - \eta\mathbf{v}^{(t,\tau)}_{k},
\end{equation}
where $\eta$ is the learning rate in global round $t$; $\mathbf{v}^{(t,\tau)}$ is the momentum in local epoch $\tau$; $\gamma \in (0,1)$ is the momentum weight. Then the local update of client $k$ takes form
\begin{equation}
    \Delta \theta^{(t)}_{k} = -\eta\sum_{\tau = 1}^{m}\mathbf{v}^{(t,\tau)}_{k}.
\end{equation}
We can compute $\mathbf{v}^{(t,\tau)}_{k}$ in Eq.~\ref{momentum_appendix} by mathematical induction, with $\mathbf{v}^{(t,1)}_{k}$ initialized as $\nabla\mathcal{L}_{\text{ce}}^{(t,1)}$. In the following, we will prove that
\begin{equation}
\label{momentum_induction}
    \mathbf{v}^{(t,\tau)}_{k} =  \sum_{i = 1}^{\tau}\gamma^{\tau - i}\nabla\mathcal{L}_{\text{ce}}^{(t,i)}.
\end{equation}
\emph{Proof: } According to Eq.~\ref{momentum_appendix},
\begin{equation}
     \mathbf{v}^{(t,2)}_{k} = \gamma \mathbf{v}^{(t,1)}_{k} + \nabla\mathcal{L}_{\text{ce}}^{(t,2)} = \gamma \nabla\mathcal{L}_{\text{ce}}^{(t,1)} + \nabla\mathcal{L}_{\text{ce}}^{(t,2)},  
\end{equation}
\begin{equation}
     \mathbf{v}^{(t,3)}_{k} = \gamma \mathbf{v}^{(t,2)}_{k} +\nabla\mathcal{L}_{\text{ce}}^{(t,3)} =  \gamma^{2} \nabla\mathcal{L}_{\text{ce}}^{(t,1)} + \gamma\nabla\mathcal{L}_{\text{ce}}^{(t,2)} + \nabla\mathcal{L}_{\text{ce}}^{(t,3)},  
\end{equation}
both satisfying Eq.~\ref{momentum_induction}. Assuming that for any $\tau \geq 2$
\begin{equation}
     \mathbf{v}^{(t,\tau)}_{k} = \sum_{i = 1}^{\tau}\gamma^{\tau - i}\nabla\mathcal{L}_{\text{ce}}^{(t,i)}, 
\end{equation}
we can obtain $\mathbf{v}^{(t,\tau+1)}_{k}$ by following Eq.~\ref{momentum_appendix},
\begin{equation}
\begin{aligned}
    \mathbf{v}^{(t,\tau+1)}_{k} &= \gamma \mathbf{v}^{(t,\tau)}_{k} + \nabla\mathcal{L}_{\text{ce}}^{(t,\tau+1)}\\
    &= \sum_{i = 1}^{\tau}\gamma^{\tau + 1 - i}\nabla\mathcal{L}_{\text{ce}}^{(t,i)} + \nabla\mathcal{L}_{\text{ce}}^{(t,\tau+1)}\\
    &= \sum_{i = 1}^{\tau+1}\gamma^{\tau + 1 - i}\nabla\mathcal{L}_{\text{ce}}^{(t,i)},
\end{aligned}
\end{equation}
which proves our claim in Eq.~\ref{momentum_induction}. Thus we have
\begin{equation}
    \Delta \theta^{(t)}_{k} = -\eta \sum_{\tau = 1}^{m} \sum_{i = 1}^{\tau}\gamma^{\tau - i}\nabla\mathcal{L}_{\text{ce}}^{(t,i)}.
\end{equation}
Similar to the discussion regarding updating bias in the output layer, we obtain
\begin{equation}
\label{sgdm_bias_appendix}
    \mathbb{E}\left[\Delta\mathbf{b}_{j}^{(t)}\right] =  \frac{\eta}{|\mathcal{B}|}\sum_{\tau = 1}^{m}\sum_{i = 1}^{\tau}\gamma^{\tau - i}\left(N_{j}^{(t,\tau)}\sum_{n \not = j} \mathcal{S}_{j,n}^{(t,\tau)} - \sum_{n \not = j} N_{n}^{(t,\tau)}\mathcal{S}_{n,j}^{(t,\tau)}\right).
\end{equation}
Reformulating Eq.~\ref{sgdm_bias_appendix} yields
\begin{equation}
     \mathbb{E}\left[\Delta\mathbf{b}_{j}^{(t)}\right] = \frac{\eta}{|\mathcal{B}|}\sum_{\tau = 1}^{m}\rho^{(\tau)}\left(N_{j}^{(t,\tau)}\sum_{n \not = j} \mathcal{S}_{j,n}^{(t,\tau)} - \sum_{n \not = j} N_{n}^{(t,\tau)}\mathcal{S}_{n,j}^{(t,\tau)}\right),
\end{equation}
where
\begin{equation}
   \rho^{(\tau)} = \frac{1-\gamma^{m + 1 - \tau}}{1-\gamma}.
\end{equation}
$\hfill \blacksquare$

\subsection{Derivation of the expected local update using Nesterov accelerated gradient method (NAG)}
\label{NAG_appendix}
In each local epoch $\tau$, Nesterov's accelerated gradient method \citep{sgd} updates the model parameters $\theta^{(t,\tau)}_{k}$ of client $k$ according to 
\begin{equation}
\label{nag_update_appendix}
    \mathbf{v}^{(t,\tau)}_{k} \xleftarrow{} \gamma \mathbf{v}^{(t,\tau-1)}_{k} + \nabla\mathcal{L}_{\text{ce}}^{(t,\tau)} + \gamma\left(\nabla\mathcal{L}_{\text{ce}}^{(t,\tau)} - \nabla\mathcal{L}_{\text{ce}}^{(t,\tau-1)}\right),
\end{equation}
\begin{equation}
    \theta^{(t,\tau)}_{k} \xleftarrow{} \theta^{(t,\tau-1)}_{k} - \eta\mathbf{v}^{(t,\tau)}_{k},
\end{equation}
where $\eta$ is the learning rate in global round $t$; $\mathbf{v}^{(t,\tau)}$ is the momentum in local epoch $\tau$; $\gamma \in (0,1)$ is the momentum weight. We form the local update of client $k$ as
\begin{equation}
    \Delta \theta^{(t)}_{k} = -\eta\sum_{\tau = 1}^{m}\mathbf{v}^{(t,\tau)}_{k}.
\end{equation}
One can compute $\mathbf{v}^{(t,\tau)}_{k}$ according to Eq.~\ref{nag_update_appendix} via mathematical induction, where $\mathbf{v}^{(t,1)}_{k}$ is initialized as $(1+\gamma)\nabla\mathcal{L}_{\text{ce}}^{(t,1)}$. In the following, we will prove that
\begin{equation}
\label{nag_induction}
     \mathbf{v}^{(t,\tau)}_{k} = \sum_{i=1}^{\tau - 1}\gamma^{\tau - i + 1}\nabla\mathcal{L}_{\text{ce}}^{(t,i)} + (1+\gamma)\nabla\mathcal{L}_{\text{ce}}^{(t,\tau)}, \tau \geq 2.
\end{equation}
\emph{Proof: } According to Eq.~\ref{nag_update_appendix},
\begin{equation}
\begin{aligned}
    \mathbf{v}^{(t,2)}_{k} &= \gamma \mathbf{v}^{(t,1)}_{k} + \nabla\mathcal{L}_{\text{ce}}^{(t,2)} + \gamma\left(\nabla\mathcal{L}_{\text{ce}}^{(t,2)} - \nabla\mathcal{L}_{\text{ce}}^{(t,1)}\right)\\
    &= \gamma^{2}\nabla\mathcal{L}_{\text{ce}}^{(t,1)} + (1+\gamma) \nabla\mathcal{L}_{\text{ce}}^{(t,2)},
\end{aligned}
\end{equation}
\begin{equation}
\begin{aligned}
    \mathbf{v}^{(t,3)}_{k} &= \gamma \mathbf{v}^{(t,2)}_{k} + \nabla\mathcal{L}_{\text{ce}}^{(t,3)} + \gamma\left(\nabla\mathcal{L}_{\text{ce}}^{(t,3)} - \nabla\mathcal{L}_{\text{ce}}^{(t,2)}\right)\\
    &= \gamma^{3}\nabla\mathcal{L}_{\text{ce}}^{(t,1)} + \gamma^{2}\nabla\mathcal{L}_{\text{ce}}^{(t,2)} + (1+\gamma)\nabla\mathcal{L}_{\text{ce}}^{(t,3)},
\end{aligned}
\end{equation}
both satisfying Eq.~\ref{nag_induction}. Assuming that for any $\tau \geq 2$
\begin{equation}
     \mathbf{v}^{(t,\tau)}_{k} = \sum_{i=1}^{\tau - 1}\gamma^{\tau - i + 1}\nabla\mathcal{L}_{\text{ce}}^{(t,i)} + (1+\gamma)\nabla\mathcal{L}_{\text{ce}}^{(t,\tau)}, 
\end{equation}
we obtain $\mathbf{v}^{(t,\tau+1)}_{k}$ according to Eq.~\ref{nag_update_appendix} as
\begin{equation}
\begin{aligned}  
    \mathbf{v}^{(t,\tau+1)}_{k} &= \gamma \mathbf{v}^{(t,\tau)}_{k} + \nabla\mathcal{L}_{\text{ce}}^{(t,\tau+1)} + \gamma\left(\nabla\mathcal{L}_{\text{ce}}^{(t,\tau+1)} - \nabla\mathcal{L}_{\text{ce}}^{(t,\tau)}\right)\\
    &= \sum_{i=1}^{\tau - 1}\gamma^{\tau+1 - i + 1}\nabla\mathcal{L}_{\text{ce}}^{(t,i)} + \gamma^{2}\nabla\mathcal{L}_{\text{ce}}^{(t,\tau)}  + (1+ \gamma)\nabla\mathcal{L}_{\text{ce}}^{(t,\tau+1)}\\
    &= \sum_{i=1}^{\tau}\gamma^{\tau+1 - i + 1}\nabla\mathcal{L}_{\text{ce}}^{(t,i)} + (1+ \gamma)\nabla\mathcal{L}_{\text{ce}}^{(t,\tau+1)},
\end{aligned}
\end{equation}
which proves our claim in Eq.~\ref{nag_induction} and thus
\begin{equation}
    \Delta \theta^{(t)}_{k} = -\eta\sum_{\tau=2}^{m}\left( \sum_{i=1}^{\tau - 1}\gamma^{\tau - i + 1}\nabla\mathcal{L}_{\text{ce}}^{(t,i)}\right) -\eta(1+\gamma)\sum_{\tau=1}^{m}\nabla\mathcal{L}_{\text{ce}}^{(t,\tau)}.
\end{equation}
Similar to the discussion regarding the update of bias in the output layer, we obtain
\begin{equation}
\label{nag_bias_appendix}
    \mathbb{E}\left[\Delta\mathbf{b}_{j}^{(t)}\right] = \frac{\eta}{|\mathcal{B}|}\sum_{\tau=1}^{m}\rho^{(\tau)}\left(N_{j}^{(t,\tau)}\sum_{n \not = j} \mathcal{S}_{j,n}^{(t,\tau)} - \sum_{n \not = j} N_{n}^{(t,\tau)}\mathcal{S}_{n,j}^{(t,\tau)}\right),
\end{equation}
where 
\begin{equation}
    \rho^{(\tau)} = \frac{1-\gamma^{m+2-\tau}}{1-\gamma}.
\end{equation}
$\hfill \blacksquare$

\subsection{Derivation of the expected local update of Scaffold}
\label{scafflod_appendix_section}
The local update of Scaffold \citep{scaffold} in local epoch $\tau$ and global round $t$ can be found as
\begin{equation}
    \theta_{k}^{(t,\tau)} \xleftarrow{} \theta_{k}^{(t,\tau-1)} - \eta\left(\nabla \mathcal{L}_{\text{ce}}^{(t,\tau)} - \mathbf{c}_{k}^{(t)} +\mathbf{c}^{(t)}\right),
\end{equation}
\begin{equation}
\label{update_rule_ck}
    \mathbf{c}_{k}^{(t+1)} \xleftarrow{} \mathbf{c}_{k}^{(t)} -  \mathbf{c}^{(t)} + \frac{1}{\eta m}(\theta^{(t)} - \theta_{k}^{(t,m)}),
\end{equation}
where $\mathbf{c}_{k}^{(t)}$ is the client control variate while $\mathbf{c}^{(t)}$ is the server control variate in global round $t$; $\mathbf{c}_{k}^{(1)}$ is initialized by $\mathbf{c}^{(1)}$; $m$ is the number of local epochs; $\eta$ is the learning rate. According to the update rule of $\mathbf{c}_{k}^{(t+1)}$ in Eq.~\ref{update_rule_ck},
\begin{equation}
\begin{aligned}
    \mathbf{c}_{k}^{(t)} &= \mathbf{c}_{k}^{(t-1)} - \mathbf{c}^{(t-1)} - \frac{1}{\eta m}\Delta \theta_{k}^{(t-1)}\\
    &= \mathbf{c}_{k}^{(t-2)} -\mathbf{c}^{(t-2)} - \mathbf{c}^{(t-1)} - \frac{1}{\eta m}\Delta \theta_{k}^{(t-2)} - \frac{1}{\eta m}\Delta \theta_{k}^{(t-1)}\\
    &= \dots\\
    &= \mathbf{c}_{k}^{(1)} -\mathbf{c}^{(1)} - \sum_{r=2}^{t-1}\mathbf{c}^{(r)} -  \frac{1}{\eta m}\sum_{r=1}^{t-1}\Delta \theta_{k}^{(r)}\\
    &=  -\sum_{r=2}^{t-1}\mathbf{c}^{(r)} -  \frac{1}{\eta m}\sum_{r=1}^{t-1}\Delta \theta_{k}^{(r)},
\end{aligned}
\end{equation}
where $\Delta \theta_{k}^{(r)}$ is the local update of client $k$ in global round $r$. Then the local update of client $k$ is given by
\begin{equation}
    \Delta \theta^{(t)}_{k} = -\eta\sum_{\tau = 1}^{m}\nabla \mathcal{L}_{\text{ce}}^{(t,\tau)} - \eta m\left(\sum_{r=2}^{t}\mathbf{c}^{(r)} +  \frac{1}{\eta m}\sum_{r=1}^{t-1}\Delta \theta_{k}^{(r)}\right).
\end{equation}
Therefore, the $j$-th component of the update of bias in the output layer can be found as
\begin{equation}
    \Delta \mathbf{b}_{j}^{(t)} = -\eta\sum_{\tau = 1}^{m}\frac{\partial \mathcal{L}_{\text{ce}}^{(t,\tau)}}{\partial\mathbf{b}_{j}}  - \eta m\left(\sum_{r=2}^{t}\mathbf{c}^{(r)} +  \frac{1}{\eta m}\sum_{r=1}^{t-1}\Delta \mathbf{b}_{j}^{(r)}\right),
\end{equation}
where $\Delta \mathbf{b}_{j}^{(r)}$ is the update of $\mathbf{b}_{j}$ in global round $r < t$ (known by the server). Similar to the previous analysis, we obtain
\begin{equation}
    \mathbb{E}\left[\Delta \mathbf{b}_{j}^{(t)}\right] = \frac{\eta}{|\mathcal{B}|}\sum_{\tau=1}^{m}\left(N_{j}^{(t,\tau)}\sum_{n \not = j} \mathcal{S}_{j,n}^{(t,\tau)} - \sum_{n \not = j} N_{n}^{(t,\tau)}\mathcal{S}_{n,j}^{(t,\tau)}\right) - \mathbf{h}_{j}^{(t)},
\end{equation}
where 
\begin{equation}
    \mathbf{h}_{j}^{(t)} = \eta m\sum_{r=2}^{t}\mathbf{c}^{(r)} +  \sum_{r=1}^{t-1}\Delta  \mathbf{b}_{j}^{(r)}.
\end{equation}
$\hfill \blacksquare$

\subsection{Derivation of the expected local update of FedProx}
\label{fedprox_appendix_section}
The local objective function in FedProx \citep{fedprox} is defined as
\begin{equation}
\label{fedprox_appendix}
    \mathcal{L}_{\text{prox}} = \mathcal{L}_{\text{ce}} + \frac{\lambda}{2}\left\Vert \theta_{k}^{(t,\tau)} - \theta^{(t)}\right\Vert^{2},
\end{equation}
where $\theta^{(t)}$ is the global model used to initialize local training; $\theta^{(t,\tau)}_{k}$ is the local model of client $k$ in local epoch $\tau$ and $\theta_{k}^{(t,1)} $ is initialized as $ \theta^{(t)}$ ; $\lambda$ is a hyper-parameter. Therefore, the gradient of $\mathcal{L}_{\text{prox}}$ in local epoch $\tau$ can be computed as
\begin{equation}
    \frac{\partial \mathcal{L}_{\text{prox}}^{(t,\tau)}}{\partial \mathbf{b}_{j}} = \frac{\partial \mathcal{L}_{\text{ce}}^{(t,\tau)}}{\partial \mathbf{b}_{j}} + \lambda\left( \mathbf{b}_{j}^{(t,\tau)} -  \mathbf{b}_{j}^{(t,1)}\right),
\end{equation}
where $\mathbf{b}_{j}^{(t,\tau)} \in \theta_{k}^{(t,\tau)}$ is the $j$-th component of bias $\mathbf{b}$ in the output layer in the local epoch $\tau$. We assume the model is trained by an SGD optimizer, leading to
\begin{equation}
\begin{aligned}
    \textcolor{blue}{\mathbf{b}_{j}^{(t,\tau)} - \mathbf{b}_{j}^{(t,1)}} &= \mathbf{b}_{j}^{(t,\tau-1)} - \eta \frac{\partial \mathcal{L}_{\text{prox}}^{(t,\tau-1)}}{\partial \mathbf{b}_{j}} - \mathbf{b}_{j}^{(t,1)} \\
    &= \mathbf{b}_{j}^{(t,\tau-1)} - \eta \frac{\partial \mathcal{L}_{\text{ce}}^{(t,\tau-1)}}{\partial \mathbf{b}_{j}} - \lambda\eta\left(\mathbf{b}_{j}^{(t,\tau-1)} -  \mathbf{b}_{j}^{(t,1)}\right) - \mathbf{b}_{j}^{(t,1)} \\
    &= -\eta \frac{\partial \mathcal{L}_{\text{ce}}^{(t,\tau-1)}}{\partial \mathbf{b}_{j}} + (1- \lambda\eta)\left(\textcolor{blue}{\mathbf{b}_{j}^{(t,\tau-1)} -  \mathbf{b}_{j}^{(t,1)}}\right)\\
    &= -\eta \frac{\partial \mathcal{L}_{\text{ce}}^{(t,\tau-1)}}{\partial \mathbf{b}_{j}} -\eta(1- \lambda\eta) \frac{\partial \mathcal{L}_{\text{ce}}^{(t,\tau-2)}}{\partial \mathbf{b}_{j}} + (1-\lambda\eta)^{2}\left(\textcolor{blue}{\mathbf{b}_{j}^{(t,\tau-2)} -  \mathbf{b}_{j}^{(t,1)}}\right)\\
    &= \dots\\
    &= -\eta\sum_{i = 1}^{\tau-1}\left(1-\lambda\eta\right)^{\tau-1-i}\frac{\partial \mathcal{L}_{\text{ce}}^{(t,i)}}{\partial \mathbf{b}_{j}} + (1-\lambda\eta)^{\tau-1}\left(\textcolor{blue}{\mathbf{b}_{j}^{(t,1)} -  \mathbf{b}_{j}^{(t,1)}}\right)\\
    &= -\eta\sum_{i = 1}^{\tau-1}\left(1-\lambda\eta\right)^{\tau-1-i}\frac{\partial \mathcal{L}_{\text{ce}}^{(t,i)}}{\partial \mathbf{b}_{j}},
\end{aligned}
\end{equation}
and thus we obtain
\begin{equation}
    \frac{\partial \mathcal{L}_{\text{prox}}^{(t,1)}}{\partial \mathbf{b}_{j}} = \frac{\partial \mathcal{L}_{\text{ce}}^{(t,1)}}{\partial \mathbf{b}_{j}}, \frac{\partial \mathcal{L}_{\text{prox}}^{(t,\tau)}}{\partial \mathbf{b}_{j}} = \frac{\partial \mathcal{L}_{\text{ce}}^{(t,\tau)}}{\partial \mathbf{b}_{j}} - \lambda\eta\sum_{i = 1}^{\tau-1}\left(1-\lambda\eta\right)^{\tau-1-i}\frac{\partial \mathcal{L}_{\text{ce}}^{(t,i)}}{\partial \mathbf{b}_{j}},\tau \geq 2.
\end{equation}
Taking expectation of both sides yields
\begin{equation}
\begin{aligned}
    \mathbb{E}\left[\Delta\mathbf{b}_{j}^{t}\right] &= -\eta\sum_{\tau=1}^{m} \mathbb{E}\left[ \frac{\partial \mathcal{L}_{\text{prox}}^{(t,\tau)}}{\partial \mathbf{b}_{j}}\right]\\
    &= -\eta\sum_{\tau=1}^{m} \mathbb{E}\left[ \frac{\partial \mathcal{L}_{\text{ce}}^{(t,\tau)}}{\partial \mathbf{b}_{j}}\right] + \lambda\eta^{2}\sum_{\tau=2}^{m}\sum_{i = 1}^{\tau-1}\left(1-\lambda\eta\right)^{\tau-1-i}\mathbb{E}\left[\frac{\partial \mathcal{L}_{\text{ce}}^{(t,i)}}{\partial \mathbf{b}_{j}}\right]\\
    &=  -\eta\sum_{\tau=1}^{m} \mathbb{E}\left[ \frac{\partial \mathcal{L}_{\text{ce}}^{(t,\tau)}}{\partial \mathbf{b}_{j}}\right] + \lambda\eta^{2}\sum_{\tau=1}^{m-1}\frac{1-(1-\lambda\eta)^{m-\tau}}{\lambda\eta}\mathbb{E}\left[\frac{\partial \mathcal{L}_{\text{ce}}^{(t,\tau)}}{\partial \mathbf{b}_{j}}\right]\\
    &= -\eta\sum_{\tau=1}^{m} \mathbb{E}\left[ \frac{\partial \mathcal{L}_{\text{ce}}^{(t,\tau)}}{\partial \mathbf{b}_{j}}\right] + \eta \sum_{\tau=1}^{m-1}\left(1-(1-\lambda\eta)^{m-\tau}\right)\mathbb{E}\left[\frac{\partial \mathcal{L}_{\text{ce}}^{(t,\tau)}}{\partial \mathbf{b}_{j}}\right],
\end{aligned}
\end{equation}
which is readily  reformulated as
\begin{equation}
    \mathbb{E}\left[\Delta\mathbf{b}_{j}^{(t)}\right] = \frac{\eta}{|\mathcal{B}|}\sum_{\tau = 1}^{m}\rho^{(\tau)}\left(N_{j}^{(t,\tau)}\sum_{n \not = j} \mathcal{S}_{j,n}^{(t,\tau)} - \sum_{n \not = j} N_{n}^{(t,\tau)}\mathcal{S}_{n,j}^{(t,\tau)}\right),
\end{equation}
where
\begin{equation}
    \rho^{(\tau)} =  (1-\lambda\eta)^{m-\tau}.
\end{equation}
$\hfill \blacksquare$
\subsection{Derivation of the expected local update of FedDyn}
\label{feddyn_appendix}
The local objective function in FedDyn \citep{feddyn} is defined as
\begin{equation}
\label{feddyn_update_appendix}
    \mathcal{L}_{\text{dyn}}^{(t,\tau)}  = \mathcal{L}_{\text{ce}} - \left\langle \nabla \mathcal{L}_{\text{ce}}^{(t-1,m)}, \theta_{k}^{(t,\tau)} \right\rangle + \frac{\lambda}{2}\left\Vert \theta_{k}^{(t,\tau)} - \theta^{(t)}\right\Vert^{2},
\end{equation}
where $\theta^{(t)}$ is the global model used to initialize local training; $\theta^{(t,\tau)}_{k}$ is the local model of client $k$ in local epoch $\tau$ and $\theta_{k}^{(t,1)} = \theta^{(t)}$ ; $\nabla \mathcal{L}_{\text{dyn}}^{(t-1,m)}$ is the gradient of $\mathcal{L}_{\text{dyn}}^{(t-1,m)}$ in the previous global round; $\lambda$ is a hyper-parameter; $m$ denotes the total number of local epochs. According to the first-order condition for local optima,
\begin{equation}
    \nabla\mathcal{L}_{\text{ce}}^{(t,\tau)}  - \nabla\mathcal{L}_{\text{ce}}^{(t-1,m)} + \lambda\left(\theta_{k}^{(t,\tau)} - \theta^{(t)}\right) = 0,
\end{equation}
and thus the partial gradient of $\mathcal{L}_{\text{ce}}$ in local epoch $m$ can be computed as
\begin{equation}
\label{feddyn_update_iteration}
\begin{aligned}
     \frac{\partial \mathcal{L}_{\text{ce}}^{(t,m)}}{\partial \mathbf{b}_{j}} &= \frac{\partial \mathcal{L}_{\text{ce}}^{(t-1,m)}}{\partial \mathbf{b}_{j}} - \lambda\left(\mathbf{b}_{j}^{(t,m)} -  \mathbf{b}_{j}^{(t,1)}\right)\\
     &= \frac{\partial \mathcal{L}_{\text{ce}}^{(t-2,m)}}{\partial \mathbf{b}_{j}} - \lambda\left( \mathbf{b}_{j}^{(t-1,m)} -  \mathbf{b}_{j}^{(t-1,1)}\right) -  \lambda\left( \mathbf{b}_{j}^{(t,m)} -  \mathbf{b}_{j}^{(t,1)}\right)\\
     &= -\lambda\sum_{r = 1}^{t}\left( \mathbf{b}_{j}^{(r,m)} -  \mathbf{b}_{j}^{(r,1)}\right)\\
     &=  -\lambda\sum_{r = 1}^{t} \Delta \mathbf{b}_{j}^{(r)},
\end{aligned}
\end{equation}
where $\Delta \mathbf{b}_{j}^{(r)}$ is the update of $\mathbf{b}_{j}$ in global round $r < t$ (known by the server); $\eta$ is the learning rate. Therefore, we obtain
\begin{equation}
     \frac{\partial \mathcal{L}_{\text{dyn}}^{(t,\tau)}}{\partial \mathbf{b}_{j}} = \frac{\partial \mathcal{L}_{\text{ce}}^{(t,\tau)}}{\partial \mathbf{b}_{j}} +\lambda\sum_{r = 1}^{t-1}\Delta \mathbf{b}_{j}^{(r)} + \lambda\left( \mathbf{b}_{j}^{(t,\tau)} - \mathbf{b}_{j}^{(t,1)} \right).
\end{equation}
Considering
\begin{equation}
\begin{aligned}
    \textcolor{blue}{\mathbf{b}_{j}^{(t,\tau)} - \mathbf{b}_{j}^{(t,1)}} &= \mathbf{b}_{j}^{(t,\tau-1)} - \eta \frac{\partial \mathcal{L}_{\text{dyn}}^{(t,\tau-1)}}{\partial \mathbf{b}_{j}} - \mathbf{b}_{j}^{(t,1)} \\
    &= \mathbf{b}_{j}^{(t,\tau-1)} - \eta \frac{\partial \mathcal{L}_{\text{ce}}^{(t,\tau-1)}}{\partial \mathbf{b}_{j}} - \lambda\eta\left(\mathbf{b}_{j}^{(t,\tau-1)} -  \mathbf{b}_{j}^{(t,1)}\right) - \mathbf{b}_{j}^{(t,1)} - \lambda \eta\sum_{r = 1}^{t-1}\Delta \mathbf{b}_{j}^{(r)}\\
    &= -\eta \frac{\partial \mathcal{L}_{\text{ce}}^{(t,\tau-1)}}{\partial \mathbf{b}_{j}} + (1- \lambda\eta)\left(\textcolor{blue}{\mathbf{b}_{j}^{(t,\tau-1)} -  \mathbf{b}_{j}^{(t,1)}}\right) - \lambda \eta \sum_{r = 1}^{t-1}\Delta \mathbf{b}_{j}^{(r)}\\
    &= -\eta \frac{\partial \mathcal{L}_{\text{ce}}^{(t,\tau-1)}}{\partial \mathbf{b}_{j}} -\eta(1- \lambda\eta) \frac{\partial \mathcal{L}_{\text{ce}}^{(t,\tau-2)}}{\partial \mathbf{b}_{j}} + (1-\lambda\eta)^{2}\left(\textcolor{blue}{\mathbf{b}_{j}^{(t,\tau-2)} -  \mathbf{b}_{j}^{(t,1)}}\right) - \lambda \eta\sum_{i=1}^{2}(1-\lambda\eta)^{2-i}\sum_{r = 1}^{t-1}\Delta \mathbf{b}_{j}^{(r)}\\
    &= \dots\\
    &= -\eta\sum_{i = 1}^{\tau-1}\left(1-\lambda\eta\right)^{\tau-1-i}\frac{\partial \mathcal{L}_{\text{ce}}^{(t,i)}}{\partial \mathbf{b}_{j}} + (1-\lambda\eta)^{\tau-1}\left(\textcolor{blue}{\mathbf{b}_{j}^{(t,1)} -  \mathbf{b}_{j}^{(t,1)}}\right) -  \lambda \eta\sum_{i=1}^{\tau-1}(1-\lambda\eta)^{\tau-1-i}\sum_{r = 1}^{t-1}\Delta \mathbf{b}_{j}^{(r)}\\
    &= -\eta\sum_{i = 1}^{\tau-1}\left(1-\lambda\eta\right)^{\tau-1-i}\frac{\partial \mathcal{L}_{\text{ce}}^{(t,i)}}{\partial \mathbf{b}_{j}} - \left(1-(1-\lambda\eta)^{\tau-1}\right)\sum_{r = 1}^{t-1}\Delta \mathbf{b}_{j}^{(r)},
\end{aligned}
\end{equation}
we obtain
\begin{equation}
\begin{aligned}
    \frac{\partial \mathcal{L}_{\text{dyn}}^{(t,\tau)}}{\partial \mathbf{b}_{j}} &= \frac{\partial \mathcal{L}_{\text{ce}}^{(t,\tau)}}{\partial \mathbf{b}_{j}} + \lambda\sum_{r = 1}^{t-1}\Delta \mathbf{b}_{j}^{(r)} - \lambda\eta\sum_{i = 1}^{\tau-1}\left(1-\lambda\eta\right)^{\tau-1-i}\frac{\partial \mathcal{L}_{\text{ce}}^{(t,i)}}{\partial \mathbf{b}_{j}}  - \lambda\left(1-(1-\lambda\eta)^{\tau-1}\right)\sum_{r = 1}^{t-1}\Delta \mathbf{b}_{j}^{(r)}\\
    &= \frac{\partial \mathcal{L}_{\text{ce}}^{(t,\tau)}}{\partial \mathbf{b}_{j}} - \lambda\eta\sum_{i = 1}^{\tau-1}\left(1-\lambda\eta\right)^{\tau-1-i}\frac{\partial \mathcal{L}_{\text{ce}}^{(t,i)}}{\partial \mathbf{b}_{j}} + \lambda(1-\lambda\eta)^{\tau-1}\sum_{r = 1}^{t-1}\Delta \mathbf{b}_{j}^{(r)},\tau \geq 2.
\end{aligned}
\end{equation}
Taking expectation of both sides yields
\begin{equation}
\begin{aligned}    \mathbb{E}\left[\Delta\mathbf{b}_{j}^{(t)}\right] &= -\eta\sum_{\tau=1}^{m} \mathbb{E}\left[ \frac{\partial \mathcal{L}_{\text{dyn}}^{(t,\tau)}}{\partial \mathbf{b}_{j}}\right]\\
    &= -\eta\sum_{\tau=1}^{m} \mathbb{E}\left[ \frac{\partial \mathcal{L}_{\text{ce}}^{(t,\tau)}}{\partial \mathbf{b}_{j}}\right] + \lambda\eta^{2}\sum_{\tau=2}^{m}\sum_{i = 1}^{\tau-1}\left(1-\lambda\eta\right)^{\tau-1-i}\mathbb{E}\left[\frac{\partial \mathcal{L}_{\text{ce}}^{(t,i)}}{\partial \mathbf{b}_{j}}\right] - \mathbf{h}^{(t)}_{j}\\
    &=  -\eta\sum_{\tau=1}^{m} \mathbb{E}\left[ \frac{\partial \mathcal{L}_{\text{ce}}^{(t,\tau)}}{\partial \mathbf{b}_{j}}\right] + \lambda\eta^{2}\sum_{\tau=1}^{m-1}\frac{1-(1-\lambda\eta)^{m-\tau}}{\lambda\eta}\mathbb{E}\left[\frac{\partial \mathcal{L}_{\text{ce}}^{(t,\tau)}}{\partial \mathbf{b}_{j}}\right]- \mathbf{h}^{(t)}_{j}\\
    &= -\eta\sum_{\tau=1}^{m} \mathbb{E}\left[ \frac{\partial \mathcal{L}_{\text{ce}}^{(t,\tau)}}{\partial \mathbf{b}_{j}}\right] + \eta \sum_{\tau=1}^{m-1}\left(1-(1-\lambda\eta)^{m-\tau}\right)\mathbb{E}\left[\frac{\partial \mathcal{L}_{\text{ce}}^{(t,\tau)}}{\partial \mathbf{b}_{j}}\right]- \mathbf{h}^{(t)}_{j},
\end{aligned}
\end{equation}
which is readily reformulated as
\begin{equation}
\mathbb{E}\left[\Delta\mathbf{b}_{j}^{(t)}\right] = \frac{\eta}{|\mathcal{B}|}\sum_{\tau = 1}^{m}\rho^{(\tau)}\left(N_{j}^{(t,\tau)}\sum_{n \not = j} \mathcal{S}_{j,n}^{(t,\tau)} - \sum_{n \not = j} N_{n}^{(t,\tau)}\mathcal{S}_{n,j}^{(t,\tau)}\right) - \mathbf{h}_{j}^{(t)},
\end{equation}
where
\begin{equation}
    \rho^{(\tau)} =  (1-\lambda\eta)^{m-\tau},
\end{equation}
\begin{equation}
    \mathbf{h}_{j}^{(t)} = \left(1 - (1-\lambda\eta)^{m}\right)\sum_{r = 1}^{t-1}\Delta \mathbf{b}_{j}^{(r)}.
\end{equation}
$\hfill \blacksquare$

\subsection{Derivation of the expected local update of FedDC}
\label{feddc_section}
The local objective function in FedDC \citep{feddc} is defined as
\begin{equation}
\label{feddc_update_appendix}
    \mathcal{L}_{\text{dc}}^{(t,\tau)}  = \mathcal{L}_{\text{ce}}+ \frac{\lambda}{2}\left\Vert \theta_{k}^{(t,\tau)} - ( \theta^{(t)} - \mathbf{d}_{k}^{(t)})\right\Vert^{2} + \frac{1}{\eta m}\left\langle \theta_{k}^{(t,\tau)}, \Delta\theta_{k}^{(t-1)} -  \Delta\theta^{(t-1)} \right\rangle,
\end{equation}
\begin{equation}
\label{feddc_update_drift}
    \mathbf{d}_{k}^{(t+1)} =  \mathbf{d}_{k}^{(t)} + \theta_{k}^{(t,m)} - \theta^{(t)},
\end{equation}
where $\mathbf{d}_{k}^{(t)}$ denotes the local drift in global round $t$ and is initialized as $\mathbf{d}_{k}^{(1)} = \mathbf{0}$; $m$ is the number of local epochs; $\Delta \theta^{(t)}$ is the global model update in round $t$. According to the update rule of $\mathbf{d}_{k}^{(t+1)}$ in Eq.~\ref{feddc_update_drift},
\begin{equation}
\begin{aligned}
    \mathbf{d}_{k}^{(t)} &= \mathbf{d}_{k}^{(t-1)} + \Delta \theta_{k}^{(t-1)}\\
    &= \mathbf{d}_{k}^{(t-2)} + \Delta \theta_{k}^{(t-2)} + \Delta \theta_{k}^{(t-1)} \\
    &= \dots\\
    &= \mathbf{d}_{k}^{(1)} + \sum_{r=1}^{t-1}\Delta \theta_{k}^{(r)}\\
    &=  \sum_{r=1}^{t-1}\Delta \theta_{k}^{(r)},
\end{aligned}
\end{equation}
and thus the gradient of $\mathcal{L}_{\text{dc}}$ in local epoch $\tau$ can be computed as
\begin{equation}
\label{feddc_update_iteration}
     \frac{\partial \mathcal{L}_{\text{dc}}^{(t,\tau)}}{\partial \mathbf{b}_{j}} = \frac{\partial \mathcal{L}_{\text{ce}}^{(t,\tau)}}{\partial \mathbf{b}_{j}} + \lambda\left(\mathbf{b}_{j}^{(t,\tau)} -  \mathbf{b}_{j}^{(t,1)} + \sum_{r=1}^{t-1}\Delta \mathbf{b}_{j}^{(r)}\right) + \frac{1}{\eta m}\left(  \Delta\mathbf{b}_{j}^{(t-1)} -  \Delta\mathbf{B}_{j}^{(t-1)}\right),
\end{equation}
where $\Delta \mathbf{B}_{j}^{(t-1)}$ denotes the global update of $\mathbf{b}_{j}$ in global round $t-1$. Then
\begin{equation}
\begin{aligned}
    \textcolor{blue}{\mathbf{b}_{j}^{(t,\tau)} - \mathbf{b}_{j}^{(t,1)}} &= \mathbf{b}_{j}^{(t,\tau-1)} - \eta \frac{\partial \mathcal{L}_{\text{dc}}^{(t,\tau-1)}}{\partial \mathbf{b}_{j}} - \mathbf{b}_{j}^{(t,1)} \\
    &= \mathbf{b}_{j}^{(t,\tau-1)} - \eta \frac{\partial \mathcal{L}_{\text{ce}}^{(t,\tau-1)}}{\partial \mathbf{b}_{j}} - \lambda\eta\left(\mathbf{b}_{j}^{(t,\tau-1)} -  \mathbf{b}_{j}^{(t,1)} + \sum_{r=1}^{t-1}\Delta \mathbf{b}_{j}^{(r)} \right) - \mathbf{b}_{j}^{(t,1)} - \frac{1}{ m}\left(  \Delta\mathbf{b}_{j}^{(t-1)} -  \Delta\mathbf{B}_{j}^{(t-1)}\right)\\
    &= -\eta \frac{\partial \mathcal{L}_{\text{ce}}^{(t,\tau-1)}}{\partial \mathbf{b}_{j}} + (1- \lambda\eta)\left(\textcolor{blue}{\mathbf{b}_{j}^{(t,\tau-1)} -  \mathbf{b}_{j}^{(t,1)}}\right) -\lambda\eta\sum_{r = 1}^{t-1}\Delta \mathbf{b}_{j}^{(r)} - \frac{1}{ m}\left(  \Delta\mathbf{b}_{j}^{(t-1)} -  \Delta\mathbf{B}_{j}^{(t-1)}\right)\\
    &= -\eta \frac{\partial \mathcal{L}_{\text{ce}}^{(t,\tau-1)}}{\partial \mathbf{b}_{j}} -\eta(1- \lambda\eta) \frac{\partial \mathcal{L}_{\text{ce}}^{(t,\tau-2)}}{\partial \mathbf{b}_{j}} + (1-\lambda\eta)^{2}\left(\textcolor{blue}{\mathbf{b}_{j}^{(t,\tau-2)} -  \mathbf{b}_{j}^{(t,1)}}\right) \\
    & \quad - \lambda\eta\sum_{i=1}^{2}(1-\lambda\eta)^{2-i}\sum_{r = 1}^{t-1}\Delta \mathbf{b}_{j}^{(r)}  - \frac{1}{m}\sum_{i=1}^{2}(1-\lambda\eta)^{2-i}\sum_{r = 1}^{t-1}\left(  \Delta\mathbf{b}_{j}^{(t-1)} -  \Delta\mathbf{B}_{j}^{(t-1)}\right) \\
    &= \dots\\
    &= -\eta\sum_{i = 1}^{\tau-1}\left(1-\lambda\eta\right)^{\tau-1-i}\frac{\partial \mathcal{L}_{\text{ce}}^{(t,i)}}{\partial \mathbf{b}_{j}} + (1-\lambda\eta)^{\tau-1}(\textcolor{blue}{\mathbf{b}_{j}^{(t,1)} -  \mathbf{b}_{j}^{(t,1)}})\\
    &\quad -  \lambda \eta\sum_{i=1}^{\tau-1}(1-\lambda\eta)^{\tau-1-i}\sum_{r = 1}^{t-1}\Delta \mathbf{b}_{j}^{(r)} - \frac{1}{m} \sum_{i=1}^{\tau-1}(1-\lambda\eta)^{\tau-1-i}\sum_{r = 1}^{t-1}\left(  \Delta\mathbf{b}_{j}^{(t-1)} -  \Delta\mathbf{B}_{j}^{(t-1)}\right)\\
    &= -\eta\sum_{i = 1}^{\tau-1}\left(1-\lambda\eta\right)^{\tau-1-i}\frac{\partial \mathcal{L}_{\text{ce}}^{(t,i)}}{\partial \mathbf{b}_{j}} - \left(1-(1-\lambda\eta)^{\tau-1}\right)\sum_{r = 1}^{t-1}\Delta \mathbf{b}_{j}^{(r)} \\
    &\quad - \frac{1}{\lambda \eta m}\left(1 - (1-\lambda\eta)^{\tau-1} \right)\left(  \Delta\mathbf{b}_{j}^{(t-1)} -  \Delta\mathbf{B}_{j}^{(t-1)}\right),
\end{aligned}
\end{equation}
and thus we obtain
\begin{equation}
\begin{aligned}
     \frac{\partial \mathcal{L}_{\text{dc}}^{(t,\tau)}}{\partial \mathbf{b}_{j}} 
     &= \frac{\partial \mathcal{L}_{\text{ce}}^{(t,\tau)}}{\partial \mathbf{b}_{j}} - \lambda\eta\sum_{i = 1}^{\tau-1}\left(1-\lambda\eta\right)^{\tau-1-i}\frac{\partial \mathcal{L}_{\text{ce}}^{(t,i)}}{\partial \mathbf{b}_{j}} + \lambda\left(1 -\lambda\eta\right)^{\tau-1}\sum_{r = 1}^{t-1}\Delta \mathbf{b}_{j}^{(r)}\\
     &\quad + \frac{(1-\lambda\eta)^{\tau-1}}{\eta m}\left(  \Delta\mathbf{b}_{j}^{(t-1)} -  \Delta\mathbf{B}_{j}^{(t-1)}\right).
\end{aligned}
\end{equation}
Taking expectation of both sides yields
\begin{equation}
\begin{aligned}    \mathbb{E}\left[\Delta\mathbf{b}_{j}^{(t)}\right] &= -\eta\sum_{\tau=1}^{m} \mathbb{E}\left[ \frac{\partial \mathcal{L}_{\text{dc}}^{(t,\tau)}}{\partial \mathbf{b}_{j}}\right]\\
    &= -\eta\sum_{\tau=1}^{m} \mathbb{E}\left[ \frac{\partial \mathcal{L}_{\text{ce}}^{(t,\tau)}}{\partial \mathbf{b}_{j}}\right] + \lambda\eta^{2}\sum_{\tau=2}^{m}\sum_{i = 1}^{\tau-1}\left(1-\lambda\eta\right)^{\tau-1-i}\mathbb{E}\left[\frac{\partial \mathcal{L}_{\text{ce}}^{(t,i)}}{\partial \mathbf{b}_{j}}\right]\\
    &\quad - \lambda\eta \sum_{\tau = 1}^{m}\left(1 -\lambda\eta\right)^{\tau-1}\sum_{r = 1}^{t-1}\Delta \mathbf{b}_{j}^{(r)} - \sum_{\tau=1}^{m}\frac{(1-\lambda\eta)^{\tau-1}}{m}\left(  \Delta\mathbf{b}_{j}^{(t-1)} -  \Delta\mathbf{B}_{j}^{(t-1)}\right)\\
    &= -\eta\sum_{\tau=1}^{m} \mathbb{E}\left[ \frac{\partial \mathcal{L}_{\text{ce}}^{(t,\tau)}}{\partial \mathbf{b}_{j}}\right] + \lambda\eta^{2}\sum_{\tau=1}^{m-1}\frac{1-(1-\lambda\eta)^{m-\tau}}{\lambda\eta}\mathbb{E}\left[\frac{\partial \mathcal{L}_{\text{ce}}^{(t,\tau)}}{\partial \mathbf{b}_{j}}\right]\\
    &\quad - \left( 1 - (1-\lambda\eta)^{m}\right)\sum_{r = 1}^{t-1}\Delta \mathbf{b}_{j}^{(r)}- \frac{1 - (1-\lambda\eta)^{m} }{\lambda\eta m}\left(  \Delta\mathbf{b}_{j}^{(t-1)} -  \Delta\mathbf{B}_{j}^{(t-1)}\right)
\end{aligned}
\end{equation}
which is readily reformulated as
\begin{equation}
\mathbb{E}\left[\Delta\mathbf{b}_{j}^{(t)}\right] = \frac{\eta}{|\mathcal{B}|}\sum_{\tau = 1}^{m}\rho^{(\tau)}\left(N_{j}^{(t,\tau)}\sum_{n \not = j} \mathcal{S}_{j,n}^{(t,\tau)} - \sum_{n \not = j} N_{n}^{(t,\tau)}\mathcal{S}_{n,j}^{(t,\tau)}\right) - \mathbf{h}_{j}^{(t)},
\end{equation}
where
\begin{equation}
    \rho^{(\tau)} =  (1-\lambda\eta)^{m-\tau},
\end{equation}
\begin{equation}
    \mathbf{h}_{j}^{(t)} = \left(1 - (1-\lambda\eta)^{m}\right)\sum_{r = 1}^{t-1}\Delta \mathbf{b}_{j}^{(r)} +  \frac{1 - (1-\lambda\eta)^{m} }{\lambda\eta m}\left(  \Delta\mathbf{b}_{j}^{(t-1)} -  \Delta\mathbf{B}_{j}^{(t-1)}\right).
\end{equation}
$\hfill \blacksquare$

\section{Experimental Details}
\subsection{Comparison with Prior Works}
\label{comparison}
\begin{table*}[t]
\caption{Comparison of RLU and state-of-the-art label recovery methods in FL. The ``Batch" column indicates whether the method can recover labels in a batch or not; ``Repeating Labels" column indicates if the method assumes no repeating labels in the batch; ``Activation-agnostic" column indicates if the method can work beyond non-negative activation functions; ``Multiple Epochs" column indicates if the method can recover labels from updates computed in multiple local epochs; ``Trained Model" column indicates if the method preserve performance on well-trained models.} 
\centering
\begin{tabular}{l|ccccc}
\bottomrule[1pt]
\label{table5}
 Schemes    & Batch & Repeating Labels &  Activation-agnostic  &  Multiple Epochs & Trained Model   \\
\hline

iDLG \citep{idlg}           & \XSolidBrush                            & \XSolidBrush                            & \XSolidBrush  &  \XSolidBrush                   &  \CheckmarkBold                     \\
GI \citep{yin}      & \CheckmarkBold                                & \XSolidBrush                              & \XSolidBrush                     &  \XSolidBrush                   & \CheckmarkBold                  \\
RLG \citep{rlg}          & \CheckmarkBold                                & \XSolidBrush                              & \CheckmarkBold &  \CheckmarkBold                   &\CheckmarkBold                      \\
LLG \citep{llg}         & \CheckmarkBold                                &\CheckmarkBold                            & \XSolidBrush                    & \XSolidBrush               &\CheckmarkBold                    \\
ZLG \cite{zlg}           & \CheckmarkBold                                & \CheckmarkBold                             & \XSolidBrush   &  \XSolidBrush                  &\XSolidBrush                     \\
iRLG \citep{irlg}          & \CheckmarkBold          & \CheckmarkBold                              & \CheckmarkBold                   &  \XSolidBrush                  &\XSolidBrush    \\
\rowcolor[HTML]{EFEFEF} 
RLU (ours)         & \CheckmarkBold          & \CheckmarkBold                             & \CheckmarkBold                   &  \CheckmarkBold                & \CheckmarkBold    \\
\toprule[1pt]
\end{tabular}
\end{table*}
We qualitatively summarize capabilities of the state-of-the-art label recovery methods in Table~\ref{table5}. A check mark (\CheckmarkBold) indicates that a method is suitable for a given setting while the cross mark (\XSolidBrush) indicates that it is not. As the table indicates, RLU is capable of operating in a variety of scenarios.

\subsection{Experimental Settings}
\label{settings_appendix}
\subsubsection{General Settings}
We used Pytorch \citep{pytorch} to implement all the described experiments. In the experiments involving SVHN, the clients used LeNet~\citep{lenet}, a convolutional neural network with $16$ convolution layers and $5 \times 5$ kernel, as the classifier. The batch size was set to $32$ in all experiments on SVHN. In the experiments involving CIFAR10 and CIFAR100, the clients trained Vgg-16 \citep{vgg} with different scales of convolution kernel as the classifier. The batch size was set to $64$ and $256$ in the experiments on CIFAR10 and CIFAR100, respectively. In the experiments involving Tiny-ImageNet, the clients learned a standard ResNet-50 \citep{resnet} where the batch size was set to $256$. The number of clients was set to $10$ across all the experiments.

\subsubsection{Settings of the Experiments in Table~\ref{table2}}
The learning rate $\eta$ for the SGD optimizer was set to $0.01$ in all experiments reported in Table~\ref{table2}. In order to obtain high level of data heterogeneity, we set the concentration parameters $\alpha$ to $0.5$ in the experiments on SVHN and CIFAR10; in the experiments on CIFAR100 and Tiny-ImageNet, this parameter was set to $0.1$. There are two groups of experiments in Table~\ref{table2}; in one the number of local epochs $m$ was set to $1$, while in the other it was set to $10$. For the experiments involving multiple local epochs, the number of iterations $T$ in Alg.~\ref{alg2} was set to $10$.

\subsubsection{Settings of the Experiments in Table~\ref{table3}}
Here we explored the effect of data heterogeneity controlled by varying the concentration parameter $\alpha$ from $0.05$ to $5$. For a fair comparison, we set the number of local epochs to $m=10$; other settings were identical to the settings in Table~\ref{table2}.

\subsubsection{Settings of the Experiments in Table~\ref{table4}, Figure~\ref{trained_models} and Figure~\ref{size_results}}
Similar to Table~\ref{table2} and Table~\ref{table3}, we set the number of local epochs $m$ to $10$ in all experiments reported in Table~\ref{table4}, Figure~\ref{trained_models} and Figure~\ref{size_results}. The concentration parameter $\alpha$ was set to $0.5$ on SVHN/CIFAR10 and to $0.1$ on CIFAR100. The learning rate $\eta$ was set to $0.01$ in experiments on SVHN and to $0.05$ in experiments on CIFAR10 and CIFAR100.

\subsubsection{Settings of  the Experiments in Gradients Inversion Attack}
We conducted gradients inversion attack on the training set of CIFAR10 with an untrained ResNet32 model. We follow the strategy in IG \citep{geiping}, using cosine similarity as the distance between ground-truth gradients and the estimated gradients computed using reconstructed images and labels. The batch size was set to $9$ and the number of iterations for optimizing the reconstructed images and labels was set to $24000$. We used Adam \citep{adam} optimizer and set the learning rate to $0.1$. We also added the total variation regularization \citep{yin} to the objective function to create more realistic images with a weight scalar $0.2$.

\subsection{Empirical Validation of Assumption \ref{assumption_1}}
\label{validation_assumption1}
\begin{figure*}[!h] 
    \centering
	  \subfloat[$\boldsymbol{\mu}_{1,1}^{(0)} = -0.14$]{
       \includegraphics[width=0.16\linewidth]{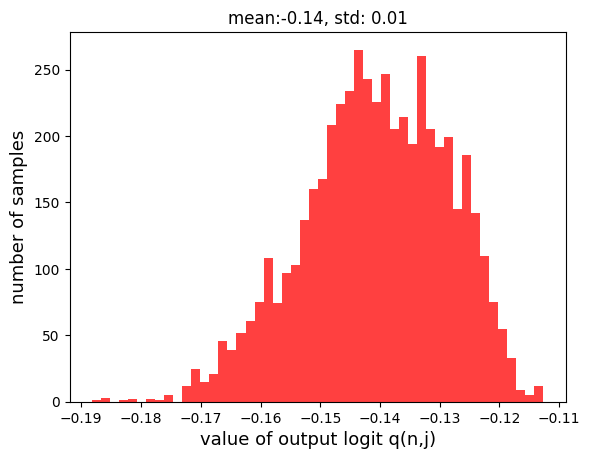}}
	  \subfloat[$\boldsymbol{\mu}_{5,5}^{(0)}  = -0.02$]{
        \includegraphics[width=0.16\linewidth]{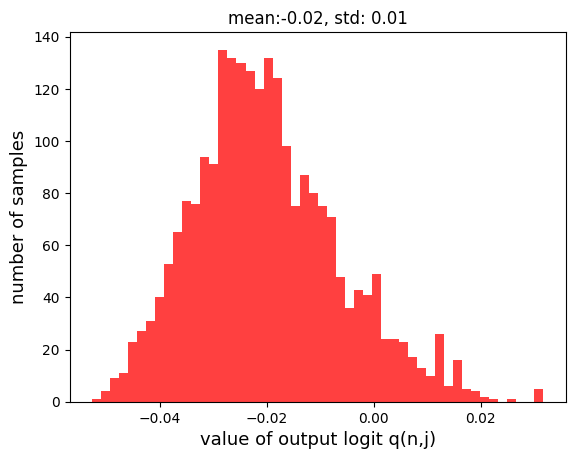}}
        \subfloat[$\boldsymbol{\mu}_{9,9}^{(0)}  = -0.04$]{
        \includegraphics[width=0.16\linewidth]{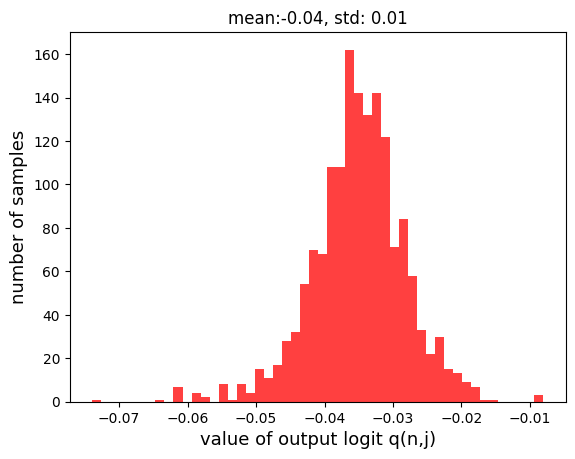}}
        \subfloat[$\boldsymbol{\mu}_{1,5}^{(0)}  = -0.02$]{
        \includegraphics[width=0.16\linewidth]{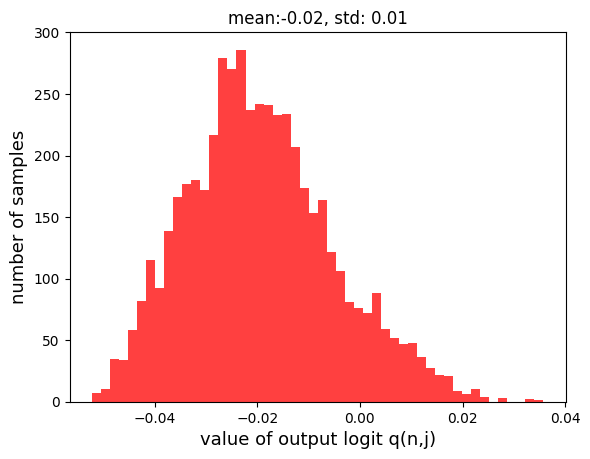}}
        \subfloat[$\boldsymbol{\mu}_{1,9}^{(0)}  = -0.04$]{
        \includegraphics[width=0.16\linewidth]{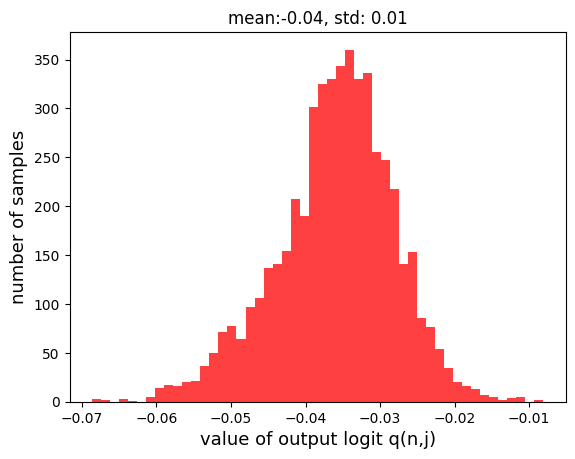}}
        \subfloat[$\boldsymbol{\mu}_{5,9}^{(0)}  = 0.02$]{
        \includegraphics[width=0.16\linewidth]{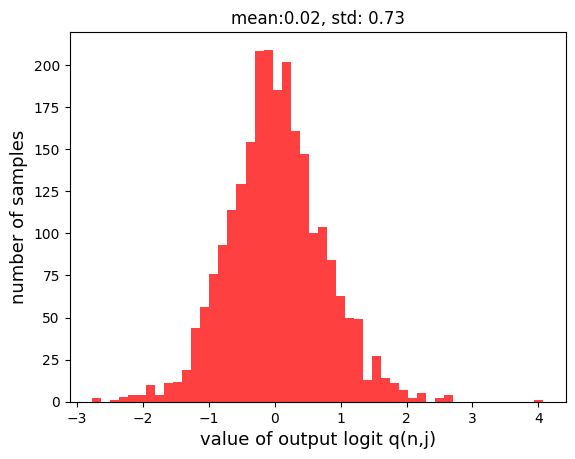}}
	\caption{A selection of histograms that characterize distribution of the output logits in global round $0$ (randomly initialized) on SVHN dataset. Specifically, each histogram illustrates the number of samples with label $n$ as input that have a certain value of the $j$-th component of output logits. From left to right, the corresponding values of $n$ and $j$ are as follow: (a) $n=1$, $j=1$; (b) $n=5$, $j=5$; (c) $n=9$, $j=9$; (d) $n=1$, $j=5$; (e) $n=1$, $j=9$; (f) $n=5$, $j=9$.}
\label{SVHN_round0} 
\end{figure*}

\begin{figure*}[!h] 
    \centering
	  \subfloat[$\boldsymbol{\mu}_{1,1}^{(4)} = 3.41$]{
       \includegraphics[width=0.16\linewidth]{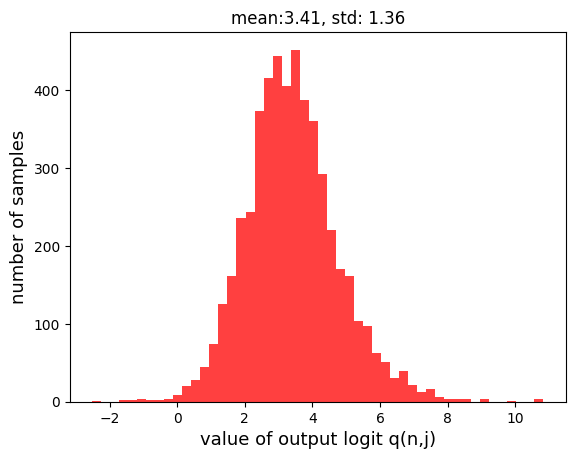}}
	  \subfloat[$\boldsymbol{\mu}_{5,5}^{(4)}  = 2.66$]{
        \includegraphics[width=0.16\linewidth]{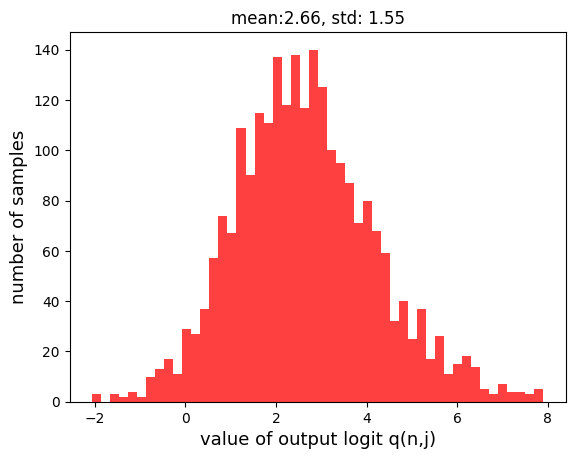}}
        \subfloat[$\boldsymbol{\mu}_{9,9}^{(4)}  = 0.89$]{
        \includegraphics[width=0.16\linewidth]{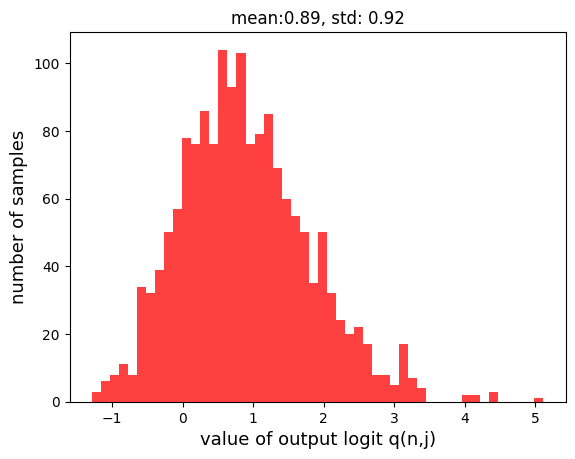}}
        \subfloat[$\boldsymbol{\mu}_{1,5}^{(4)}  = -0.96$]{
        \includegraphics[width=0.16\linewidth]{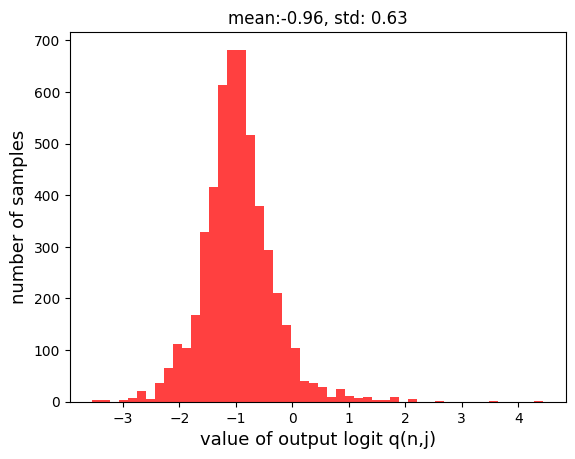}}
        \subfloat[$\boldsymbol{\mu}_{1,9}^{(4)}  = -1.29$]{
        \includegraphics[width=0.16\linewidth]{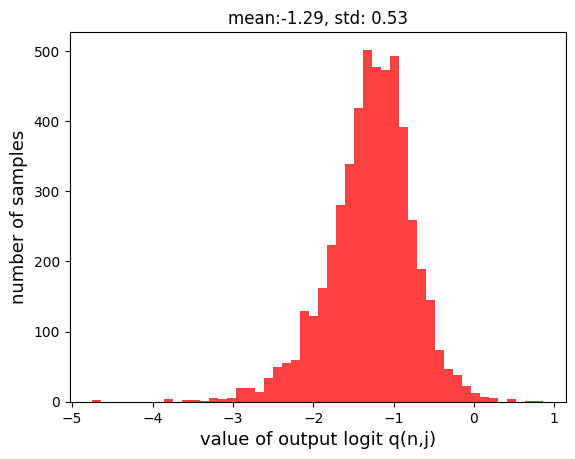}}
        \subfloat[$\boldsymbol{\mu}_{5,9}^{(4)}  = -0.03$]{
        \includegraphics[width=0.16\linewidth]{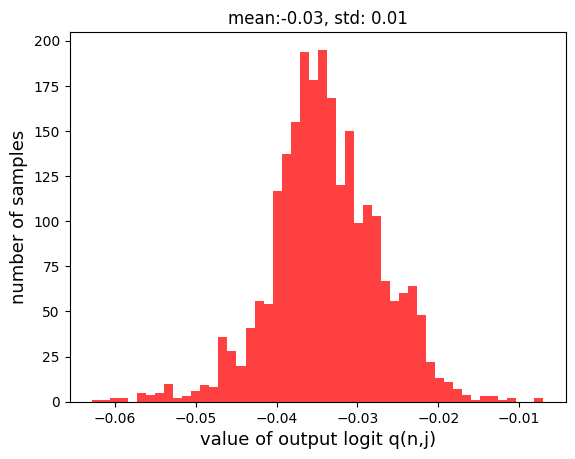}}
	\caption{A selection of histograms that characterize  distribution of the output logits in global round $4$ in which the training accuracy of the global model is $68\%$ on SVHN dataset. The other settings are identical to Fig. \ref{SVHN_round0}.}
\label{SVHN_round4} 
\end{figure*}

\begin{figure*}[!h] 
    \centering
	  \subfloat[$\boldsymbol{\mu}_{1,1}^{(9)} = 4.92$]{
       \includegraphics[width=0.16\linewidth]{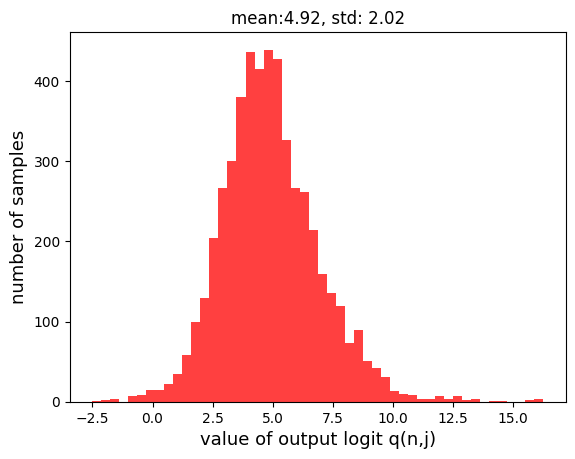}}
	  \subfloat[$\boldsymbol{\mu}_{5,5}^{(9)}  = 5.65$]{
        \includegraphics[width=0.16\linewidth]{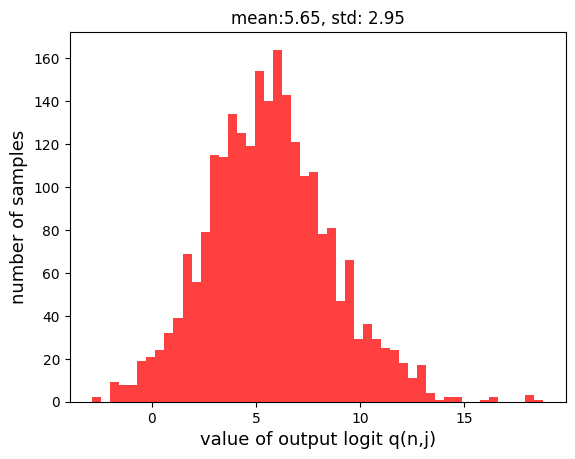}}
        \subfloat[$\boldsymbol{\mu}_{9,9}^{(9)}  = 3.38$]{
        \includegraphics[width=0.16\linewidth]{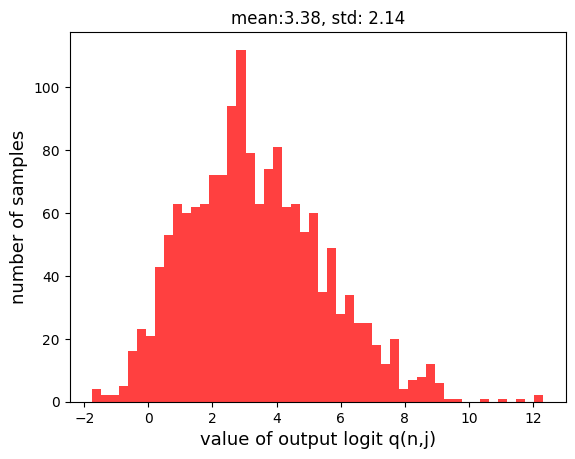}}
        \subfloat[$\boldsymbol{\mu}_{1,5}^{(9)}  = -1.51$]{
        \includegraphics[width=0.16\linewidth]{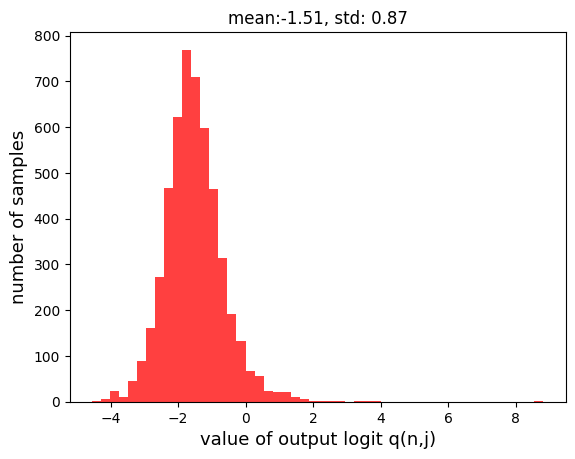}}
        \subfloat[$\boldsymbol{\mu}_{1,9}^{(9)}  = -1.39$]{
        \includegraphics[width=0.16\linewidth]{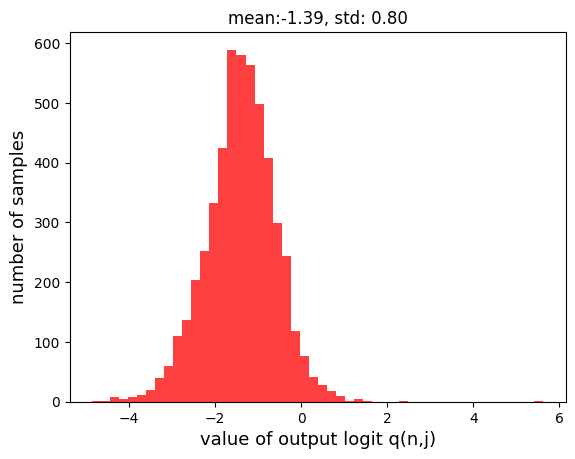}}
        \subfloat[$\boldsymbol{\mu}_{5,9}^{(9)}  = -0.56$]{
        \includegraphics[width=0.16\linewidth]{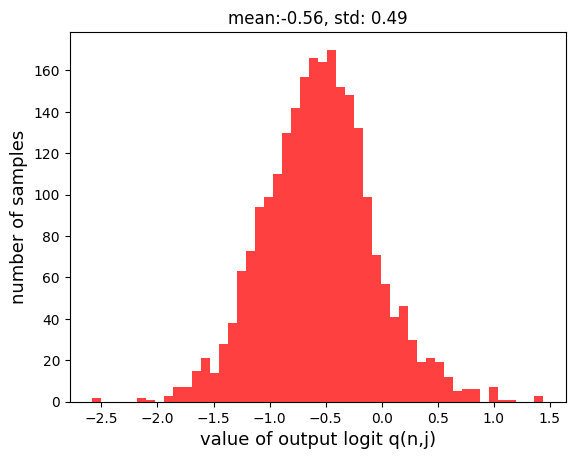}}
	\caption{A selection of histograms that characterize  distribution of the output logits in the global round $9$ in which the training accuracy of the global model is $83\%$ on SVHN dataset. The other settings are identical to Fig. \ref{SVHN_round0}.}
\label{SVHN_round9} 
\end{figure*}

\begin{figure*}[!h] 
    \centering
	  \subfloat[$\boldsymbol{\mu}_{1,1}^{(0)} = 0.06$]{
       \includegraphics[width=0.16\linewidth]{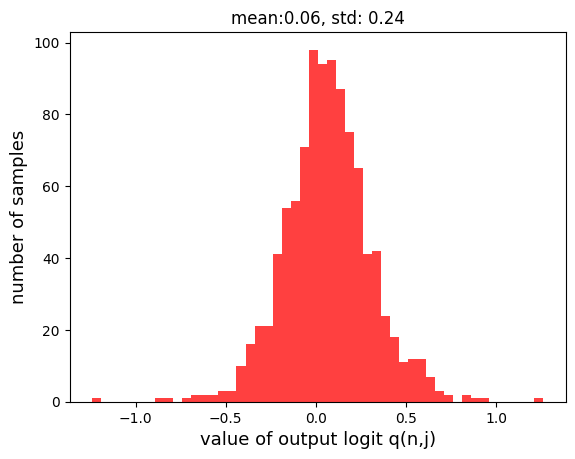}}
	  \subfloat[$\boldsymbol{\mu}_{5,5}^{(0)}  = 0.03$]{
        \includegraphics[width=0.16\linewidth]{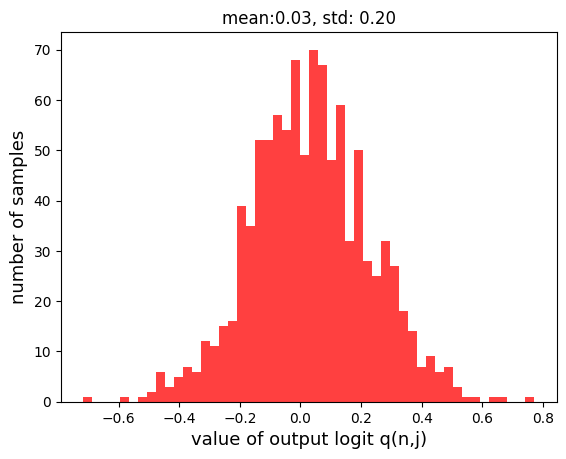}}
        \subfloat[$\boldsymbol{\mu}_{9,9}^{(0)}  = 0.20$]{
        \includegraphics[width=0.16\linewidth]{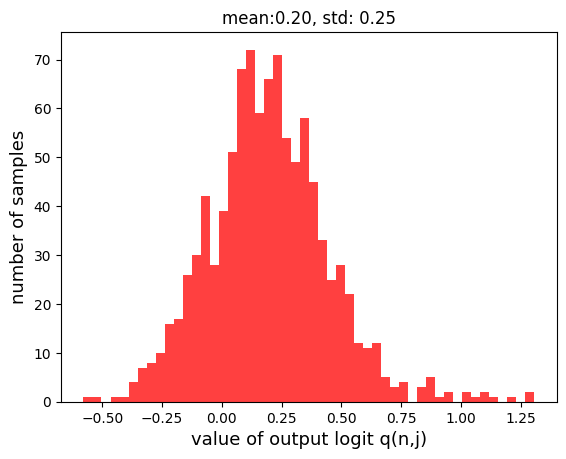}}
        \subfloat[$\boldsymbol{\mu}_{1,5}^{(0)}  = 0.06$]{
        \includegraphics[width=0.16\linewidth]{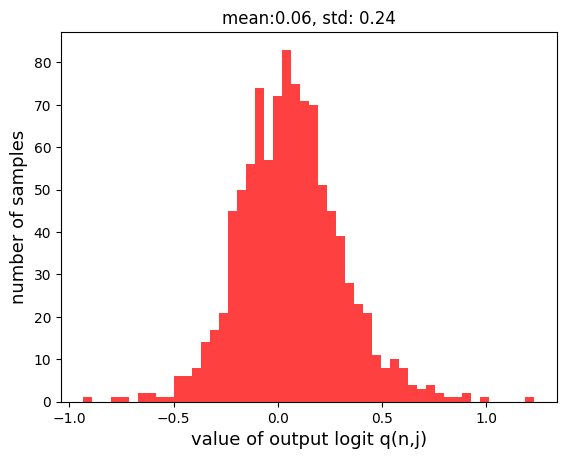}}
        \subfloat[$\boldsymbol{\mu}_{1,9}^{(0)}  = 0.21$]{
        \includegraphics[width=0.16\linewidth]{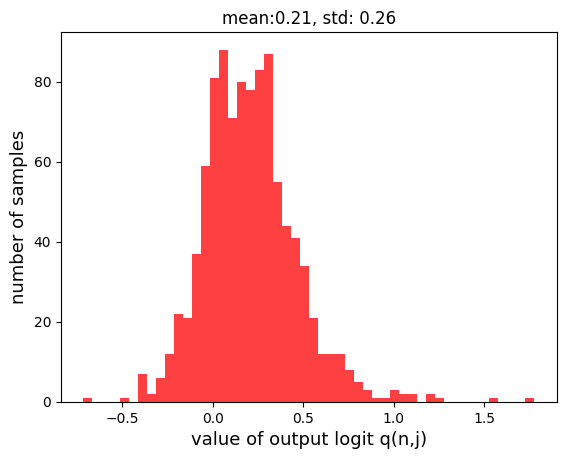}}
        \subfloat[$\boldsymbol{\mu}_{5,9}^{(0)}  = 0.18$]{
        \includegraphics[width=0.16\linewidth]{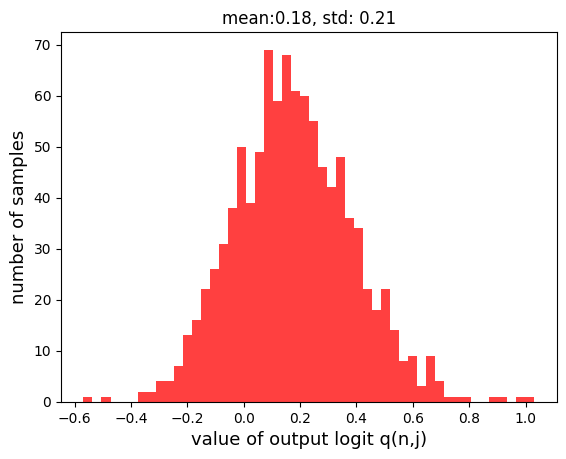}}
	\caption{A selection of histograms that characterize  distribution of the output logits in global round $0$ on CIFAR10. The other settings are identical to Fig. \ref{SVHN_round0}.}
\label{CIFAR10_round0} 
\end{figure*}

\begin{figure*}[!h] 
    \centering
	  \subfloat[$\boldsymbol{\mu}_{1,1}^{(20)} = 6.21$]{
       \includegraphics[width=0.16\linewidth]{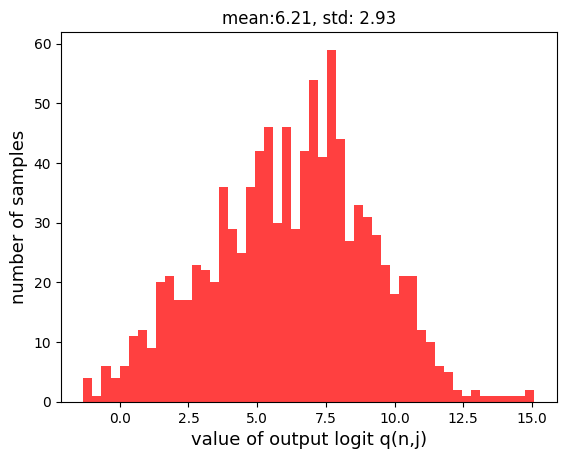}}
	  \subfloat[$\boldsymbol{\mu}_{5,5}^{(20)}  = 4.56$]{
        \includegraphics[width=0.16\linewidth]{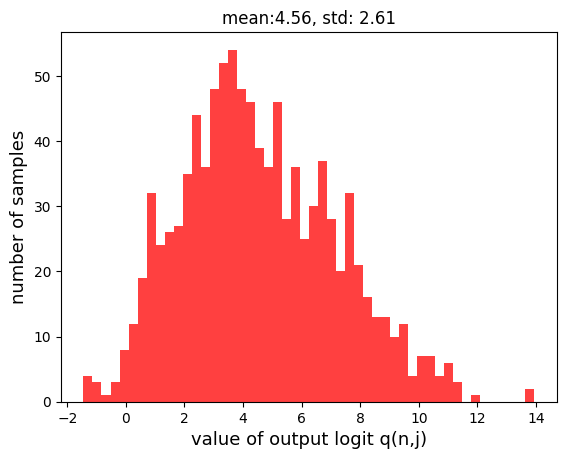}}
        \subfloat[$\boldsymbol{\mu}_{9,9}^{(20)}  = 3.46$]{
        \includegraphics[width=0.16\linewidth]{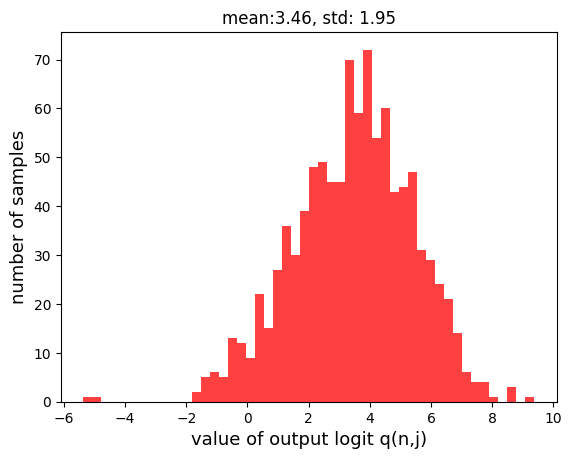}}
        \subfloat[$\boldsymbol{\mu}_{1,5}^{(20)}  = -1.83$]{
        \includegraphics[width=0.16\linewidth]{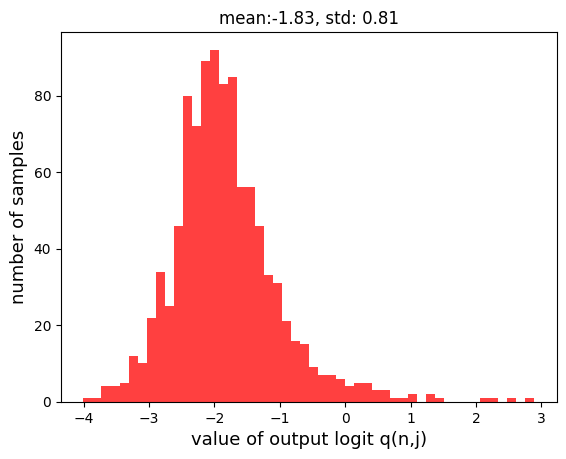}}
        \subfloat[$\boldsymbol{\mu}_{1,9}^{(20)}  = 2.04$]{
        \includegraphics[width=0.16\linewidth]{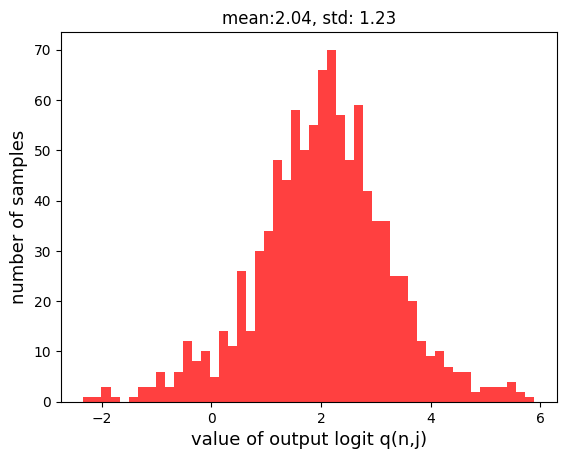}}
        \subfloat[$\boldsymbol{\mu}_{5,9}^{(20)}  = -2.31$]{
        \includegraphics[width=0.16\linewidth]{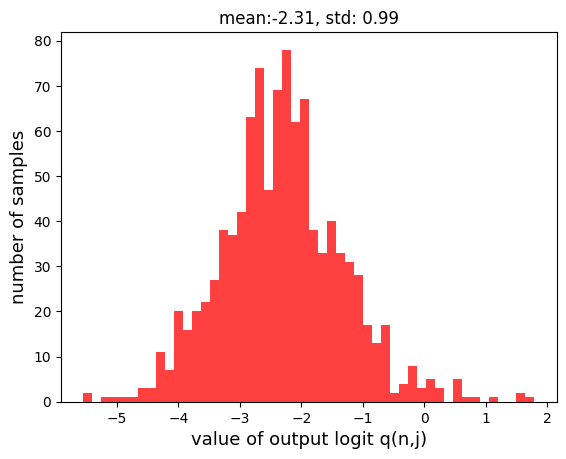}}
	\caption{A part of histograms that characterize the distribution of output logits in the global round $20$ in which the training accuracy of the global model is $80\%$ on CIFAR10. The other settings are identical to Fig. \ref{SVHN_round0}.}
\label{CIFAR10_round20} 
\end{figure*}

\begin{figure*}[!h] 
    \centering
	  \subfloat[$\boldsymbol{\mu}_{1,1}^{(40)} = 8.46$]{
       \includegraphics[width=0.16\linewidth]{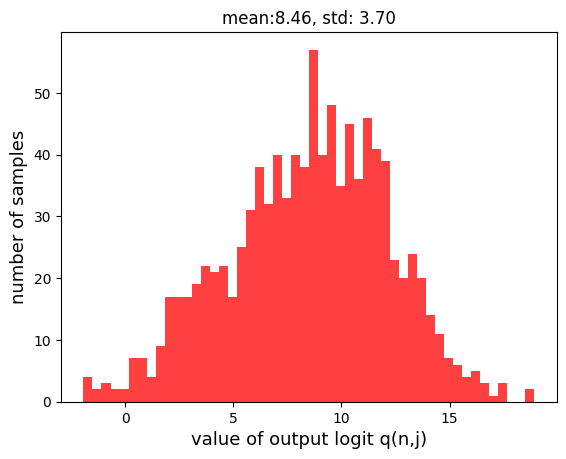}}
	  \subfloat[$\boldsymbol{\mu}_{5,5}^{(40)}  = 6.78$]{
        \includegraphics[width=0.16\linewidth]{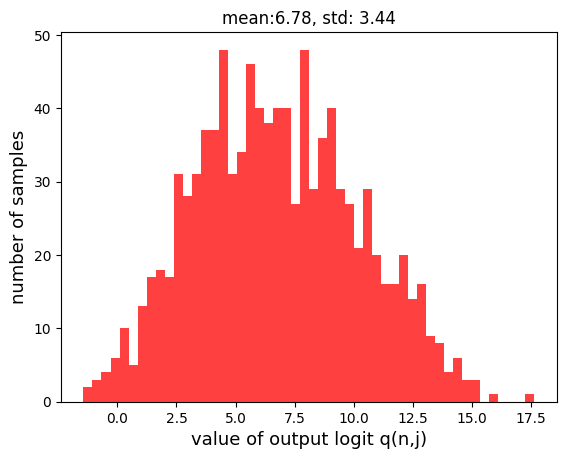}}
        \subfloat[$\boldsymbol{\mu}_{9,9}^{(40)}  = 5.36$]{
        \includegraphics[width=0.16\linewidth]{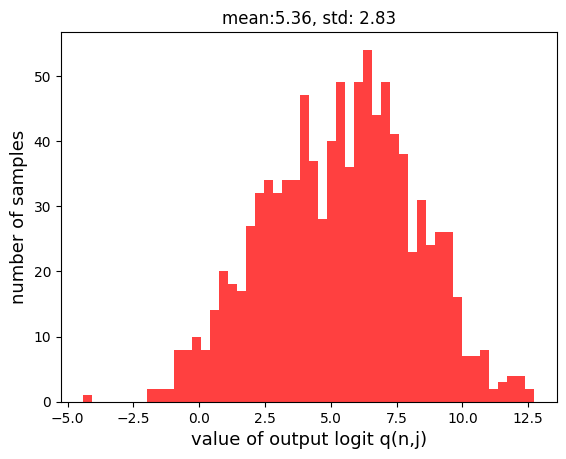}}
        \subfloat[$\boldsymbol{\mu}_{1,5}^{(40)}  = -1.97$]{
        \includegraphics[width=0.16\linewidth]{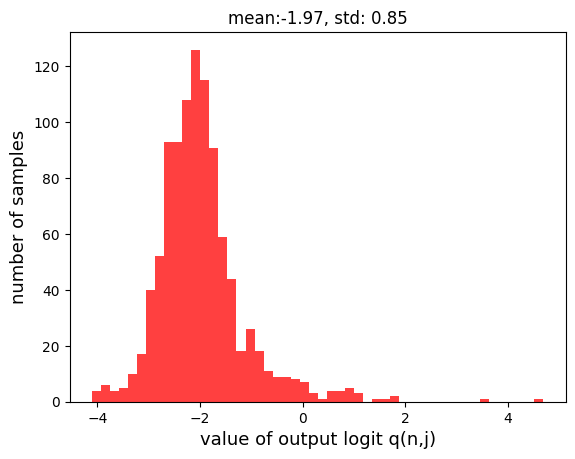}}
        \subfloat[$\boldsymbol{\mu}_{1,9}^{(40)}  = 2.21$]{
        \includegraphics[width=0.16\linewidth]{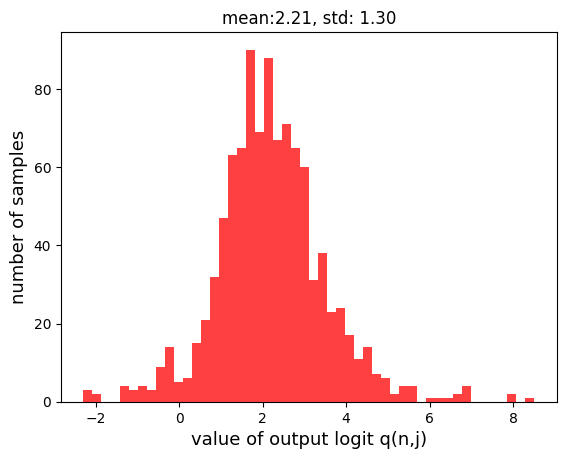}}
        \subfloat[$\boldsymbol{\mu}_{5,9}^{(40)}  = -2.67$]{
        \includegraphics[width=0.16\linewidth]{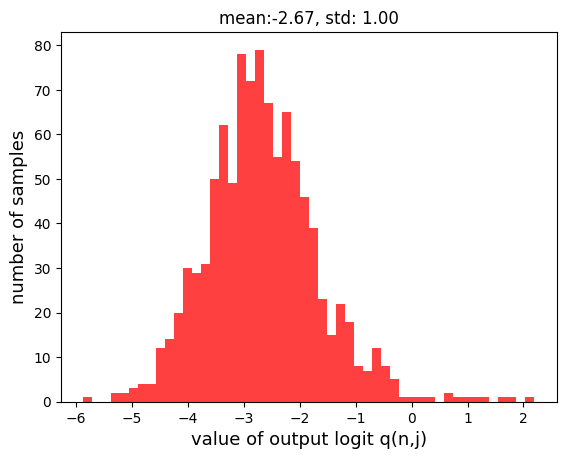}}
	\caption{A part of histograms that characterize the distribution of output logits in the global round $40$ in which the training accuracy of the global model is $95\%$ on CIFAR10. The other settings are identical to Fig. \ref{SVHN_round0}.}
\label{CIFAR10_round40} 
\end{figure*}

To verify Assumption \ref{assumption_1}, we perform inference on the training set of SVHN and CIFAR10 using the global model across different training rounds. We collect the $j$-th component of the output logits for the samples with label $n$, and organize them into $100$ bins to generate histograms. In each histogram, x-axis indicates the values of the $j$-th component of the output logits while the values on y-axis indicate the corresponding number of samples. As shown in Figures~\ref{SVHN_round0} and \ref{CIFAR10_round0}, means $\boldsymbol{\mu}_{n,j}$ of the randomly initialized global model are close to $0$ regardless of their sign. As the training accuracy of the global model improves, the means $\boldsymbol{\mu}_{j,j}$ increase to larger positive values while $\boldsymbol{\mu}_{n,j} (n \not = j)$ decreases to negative values, as shown in Figures \ref{SVHN_round0}-\ref{CIFAR10_round40}.

\subsection{Evaluation Metrics}
\label{metrics}
We follow the strategy in iRLG \citep{irlg}, evaluating all methods in terms of class-level accuracy (cAcc) and instance-level accuracy (iAcc). Specifically, cAcc indicates the proportion of correctly recovered
classes while iAcc indicates the proportion of correctly recovered
labels. Suppose $\mathbf{y}_{c}$ denotes the ground-truth classes that appear in the batches while $\hat{\mathbf{y}}_{c}$ denotes the estimated classes; then cAcc can be computed as
\begin{equation}
    \textbf{cAcc}(\hat{\mathbf{y}}_{c}, \mathbf{y}_{c}) = \frac{|\hat{\mathbf{y}}_{c} \textbf{ XNOR } \mathbf{y}_{c}|}{N},
\end{equation}
where $N$ is the total number of classes. Similarly, suppose  $\mathbf{y}_{i}$ denotes the ground-truth labels in the batches while $\hat{\mathbf{y}}_{i}$ denotes estimated labels; then iAcc can be computed as
\begin{equation}
    \textbf{iAcc}(\hat{\mathbf{y}}_{i}, \mathbf{y}_{i}) = \frac{|\hat{\mathbf{y}}_{i} \cap \mathbf{y}_{i}|}{m \cdot |\mathcal{B}|},
\end{equation}
where $|\hat{\mathbf{y}}_{i}| = |\mathbf{y}_{i}| = m \cdot |\mathcal{B}|$ and $m$ denotes the number of local epochs.

\subsection{Visualization of Data Heterogeneity}
\label{visualization}
\begin{figure*}[h] 
    \centering
	  \subfloat[$\alpha = 0.05$]{
       \includegraphics[width=0.25\linewidth]{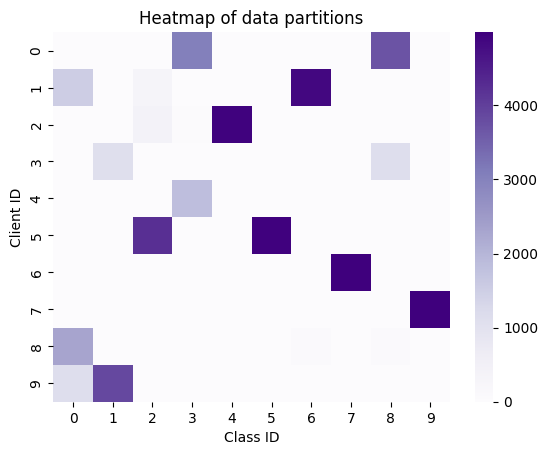}}
	  \subfloat[$\alpha = 0.5$]{
        \includegraphics[width=0.25\linewidth]{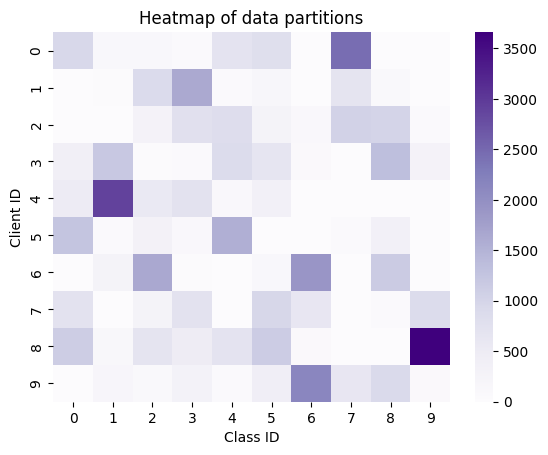}}
        \subfloat[$\alpha = 5$]{
        \includegraphics[width=0.25\linewidth]{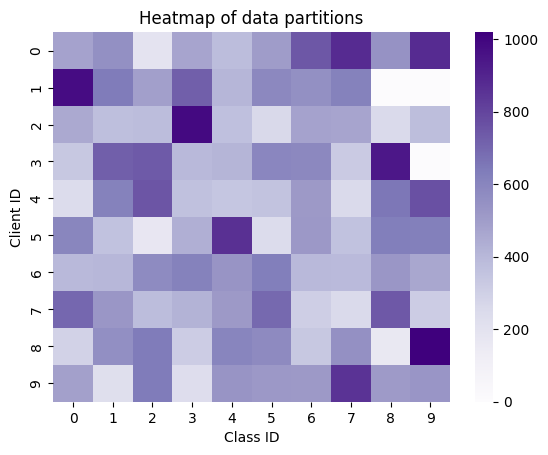}}
	\caption{Training data from CIFAR10 is split into 10 partitions according to a Dirichlet distribution. The concentration parameter is set as follows: (a) $\alpha = 0.05$; (b) $\alpha = 0.5$; (c) $\alpha = 5$.}
\label{cifar10_visualization} 
\end{figure*}

\begin{figure*}[h] 
    \centering
	  \subfloat[$\alpha = 0.05$]{
       \includegraphics[width=0.25\linewidth]{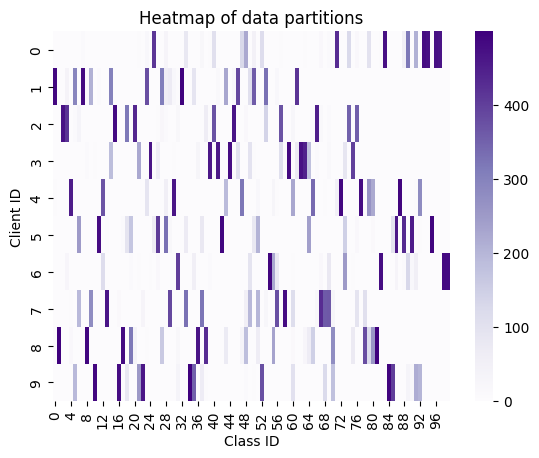}}
	  \subfloat[$\alpha = 0.5$]{
        \includegraphics[width=0.25\linewidth]{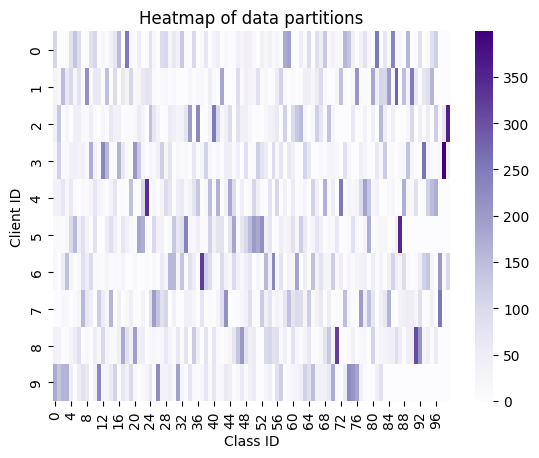}}
        \subfloat[$\alpha = 5$]{
        \includegraphics[width=0.25\linewidth]{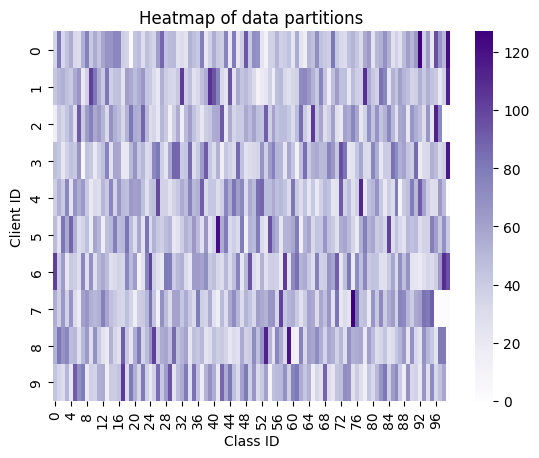}}
	\caption{Training data from CIFAR100 is split into 10 partitions according to a Dirichlet distribution. The concentration parameter is set as follows: (a) $\alpha = 0.05$; (b) $\alpha = 0.5$; (c) $\alpha = 5$.}
\label{cifar100_visualization} 
\end{figure*}

We follow the strategy in \citep{bayesian}, utilizing Dirichlet distribution to generate heterogeneous data partitions from the original datasets. Specifically, we assign different proportions $\mathbf{p}_{k}^{(j)}$ of samples with label $j$ in the local dataset $\mathcal{D}_{k}$ to $K$ clients according to
\begin{equation}
    \mathbf{p}^{(j)} = \{\mathbf{p}_{k}^{(j)}, k \in [K]\} \sim \textbf{Dir}_{K}(\alpha),
\end{equation}
where $\alpha$ is the concentration parameter that controls the level of heterogeneity. The number of samples with label $j$ in client $k$'s local dataset can be computed as 
\begin{equation}
    N_{k}^{(j)} = \frac{\mathbf{p}_{k}^{(j)}}{\sum_{i}^{K}\mathbf{p}_{i}^{(j)}} N^{(j)},
\end{equation}
where $N^{(j)}$ is the number of samples with label $j$ in the overall training dataset. Figures \ref{cifar10_visualization} and \ref{cifar100_visualization} show the class distribution of clients' local dataset by color-coding the number of samples: the darker the color, the larger the number of samples with the corresponding label. As shown in Figures \ref{cifar10_visualization} and \ref{cifar100_visualization}, clients own only 2 or 3 classes in their local dataset given $\alpha = 0.05$ while $\alpha = 5$ leads to more balanced class distribution.

\section{Algorithms}
\subsection{RLU Algorithm}
\label{rlu}
In this section, we formalize the RLU algorithm as Alg.\ref{alg1}. In the standard paradigm in federated learning, the server broadcasts the global model $\boldsymbol{\theta}^{(t)}$ to selected clients; these clients then initialize their local model $\boldsymbol{\theta}_{k}^{(t)}$ to start local training. After $m$ local epochs, each client $k$ computes the local update $\Delta \boldsymbol{\theta}_{k}^{(t)} = \boldsymbol{\theta}_{k}^{(t)} - \boldsymbol{\theta}^{(t)}$ including $\Delta \mathbf{W}_{k}^{(t)}$ and $\Delta \mathbf{b}_{k}^{(t)}$. For convenience, we omit the subscript $k$ in $\Delta \mathbf{W}_{k}^{(t)}$ and $\Delta \mathbf{b}_{k}^{(t)}$ in Alg.\ref{alg1}. According to the formulated problem in Section \ref{problem_formulation}, the learning rate $\eta$ in local training is known by the server. While preparing to conduct label recovery attack, the server estimates $\boldsymbol{\mu}_{n}^{(t)}, \Sigma_{n}^{(t)}, \mathcal{S}_{n,j}^{(t)},\boldsymbol{\mu}_{n,j}^{(t+1)}, \Sigma_{n}^{(t+1)}$, and $\mathcal{S}_{n,j}^{(t+1)}, \forall n,j \in [N]$ via Monte Carlo method according to Eq.~\ref{monte_carlo} with the global model $\boldsymbol{\theta}^{(t)}$, updated local model $\boldsymbol{\theta}^{(t)}_{k}$ and the auxiliary dataset $\mathcal{A}$. If $m = 1$, the server simply solves the least-squares problem described in Eq.~\ref{minimization}. If $m > 1$, the server first solves the least-squares problem using coefficients computed as the mean of $\mathcal{S}_{n,j}^{(t)}$ and $\mathcal{S}_{n,j}^{(t+1)}$ to obtain a crude estimate. Then the server conducts posterior search (Alg.~\ref{alg2}) to adjust the estimated $N^{(t)}$ as discussed in Section \ref{multiple_epochs}.
\begin{algorithm}[!h]
  \KwIn{Auxiliary dataset $\mathcal{A}$, the global model $\boldsymbol{\theta}^{(t)}$, local updates $\Delta \boldsymbol{\theta}^{(t)}_{k}$ including $\Delta \mathbf{W}^{(t)}$ and $\Delta \mathbf{b}^{(t)}$, learning rate $\eta$.}
  \KwOut{The number of samples with each label $N_{j}^{(t)}$ in the batches sampled by client $k$, where $\forall j \in [N]$.}
  \textbf{Initial}: Estimate $\boldsymbol{\mu}_{n}^{(t)}, \boldsymbol{\Sigma}_{n}^{(t)}, \mathcal{S}_{n,j}^{(t)},\boldsymbol{\mu}_{n,j}^{(t+1)}, \boldsymbol{\Sigma}_{n}^{(t+1)}$, and $\mathcal{S}_{n,j}^{(t+1)}, \forall n,j \in [N]$ via Monte Carlo method according to Eq.\ref{monte_carlo} with the global model $\boldsymbol{\theta}^{(t)}$, updated local model $\boldsymbol{\theta}^{(t)}_{k} \xleftarrow{} \boldsymbol{\theta}^{(t)} + \Delta \boldsymbol{\theta}^{(t)}_{k}$ and the auxiliary dataset $\mathcal{A}$; create coefficient matrices $\mathbf{A}^{(t)}, \mathbf{A}^{(t+1)}  \in \mathbb{R}^{N \times N}$; initialize vector $\mathbf{u} \xleftarrow{} \Delta \mathbf{b}^{(t)}/\eta$.\;
  
  \tcc{\textcolor{blue}{Compute coefficients for the least square problem}}
  
  \For{$n, j \in [N]$}{
     \eIf{$n \not = j$}{$\mathbf{A}_{n,j}^{(t)} \xleftarrow{} -\mathcal{S}_{n,j}^{(t)}$, $\mathbf{A}_{n,j}^{(t+1)} \xleftarrow{} -\mathcal{S}_{n,j}^{(t+1)}$\;}
     {$\mathbf{A}_{j,j}^{(t)} \xleftarrow{} \sum_{n\not = j}\mathcal{S}_{j,n}^{(t)} $, $\mathbf{A}_{j,j}^{(t+1)} \xleftarrow{} \sum_{n\not = j}\mathcal{S}_{j,n}^{(t+1)} $ \;}
  }   
  \eIf{$m = 1$}{
        $\mathbf{z}  \xleftarrow{} \textbf{LeastSquare}(\mathbf{A}^{(t)}, \mathbf{u})$ as described in (\ref{minimization})\;

        $\mathbf{N}^{(t)} \xleftarrow{} \lfloor|\mathcal{B}|\cdot \mathbf{z}\rceil$\;
        }
        {
      $\bar{\mathbf{A}} \xleftarrow{} (\mathbf{A}^{(t)} + \mathbf{A}^{(t+1)})/2$\;

       $\mathbf{z} \xleftarrow{} \textbf{LeastSquare}(\bar{\mathbf{A}}, \mathbf{u})$, $\mathbf{N}^{(t)} \xleftarrow{} \lfloor|\mathcal{B}|\cdot \mathbf{z}\rceil$\;

        $\mathbf{N}^{(t)}  \xleftarrow{} \textbf{PosteriorSearch}(\mathbf{N}^{(t)}, \Delta \mathbf{W}^{(t)}, \Delta \mathbf{b}^{(t)}, \boldsymbol{\mu}_{n}^{(t)},  \boldsymbol{\Sigma}_{n}^{(t)}, \boldsymbol{\mu}_{n}^{(t+1)},\boldsymbol{\Sigma}_{n}^{(t+1)}, \eta)$\;
      }
    \Return $\mathbf{N}^{(t)}: \{N_{j}^{(t)}, \forall j \in [N]\}$
  \caption{RLU}\label{alg1}
\end{algorithm}

\subsection{Posterior Search Algorithm}
\label{search}
In Alg.~\ref{alg2}, we simulated the dynamics of $\boldsymbol{\mu}_{n}^{(t)}$ during local training. First, the server estimates average embedded signal $\Bar{\mathbf{e}}$ according to Eq.~\ref{expectation_of_bias_multiple}. Using the crude estimates of $N^{(t)}$ and $\mathcal{S}_{n,j}^{(t)}$ as inputs, the server computes $\Delta \mathbf{b}_{j}^{(t,\tau)}$ according to Eq.~\ref{update_of_bias}. Subsequently, the server computes $\Delta \boldsymbol{\mu}_{n,j}^{(t,\tau)}$ using $\Bar{\mathbf{e}}$, and then estimates $\hat{\boldsymbol{\mu}}_{n, j}^{(t,\tau+1)}$. Recursively, repeating the procedure, the server finally obtains $\hat{\boldsymbol{\mu}}_{n, j}^{(t,\tau+1)}$. Based on the 
difference between $\hat{\boldsymbol{\mu}}_{n, j}^{(t,\tau+1)}$ and $\boldsymbol{\mu}_{n, j}^{(t,\tau+1)}$, the server can calibrate/improve estimated $N^{(t)}$. 
\begin{algorithm}
  \KwIn{Number of iterations $T$, $\Delta \mathbf{W}^{(t)}$, $\Delta \mathbf{b}^{(t)}$, learning rate $\eta$, estimated $\mathbf{N}^{(t)}$, $\boldsymbol{\mu}_{n}^{(t)},\boldsymbol{\Sigma}_{n}^{(t)},\boldsymbol{\mu}_{n}^{(t+1)}$ and $\boldsymbol{\Sigma}_{n}^{(t+1)}$ for $n,j \in [N]$.}
  \KwOut{The number of samples with each label $N_{j}^{(t)}$ in the batches, where $\forall j \in [N]$.}
  \textbf{Initial}: The average embedding signal $\bar{\mathbf{e}} \in \mathbb{R}^{L}$. \;
  
  \tcc{\textcolor{blue}{Compute average embedding signal.}}
  
  \For{$l \in [L]$}{
      $\bar{\mathbf{e}}_{l} \xleftarrow{} \Delta \mathbf{W}^{(t)}_{j,l}/ \Delta \mathbf{b}^{(t)}_{j}$\;
  }   
 \For{$n \in [N]$}{
      $\hat{\boldsymbol{\mu}}_{n}^{(t,1)} \xleftarrow{} \boldsymbol{\mu}_{n}^{(t)}$\;
  }  
     \For{$i \in [T]$}{
     $\mathbf{g} \xleftarrow{} \lfloor \mathbf{N}^{(t)}/m\rceil$\;
     
        \For{$\tau \in [m]$}{
         \tcc{\textcolor{blue}{Update the expectation}}
        $\hat{\mathcal{S}}_{n,j}^{(t,\tau)} \xleftarrow{} \textbf{MonteCarlo}(\hat{\boldsymbol{\mu}}_{n}^{(t,\tau)},\boldsymbol{\Sigma}_{n}^{(t)})$, for $n,j \in [N]$\;
        
            \For{$j \in [N]$}{
            \tcc{\textcolor{blue}{Use the statics of model to estimate the updates of bias as illustrated in Eq.\ref{update_of_bias}.}}
             $\Delta \mathbf{b}_{j}^{(t,\tau)} \xleftarrow{} \frac{\eta}{|\mathcal{B}|}\left(\mathbf{g}_{j}\sum_{n \not= j}^{N} \hat{\mathcal{S}}_{j,n}^{(t,\tau)} - \sum_{n \not = j} \mathbf{g}_{n}\hat{\mathcal{S}}_{n,j}^{(t,\tau)}\right)$\;
            }
            \For{$j \in [N]$}{
            \tcc{\textcolor{blue}{Update the intermediate means according to Eq.\ref{use_signal_to_update}.}}
             $\Delta \boldsymbol{\mu}_{n, j}^{(t,\tau)} \xleftarrow{} \Delta \mathbf{b}_{j}^{(t,\tau)} \cdot \sum_{l}^{L} \bar{\mathbf{e}}_{l}^{2}$, for $ \forall n \in [N]$\;
    
             $\hat{\boldsymbol{\mu}}_{n, j}^{(t,\tau+1)} \xleftarrow{} \hat{\boldsymbol{\mu}}_{n, j}^{(t,\tau)} + \Delta \boldsymbol{\mu}_{n, j}^{(t,\tau)} $, for $\forall n \in [N]$\;
            }
        }
        \tcc{\textcolor{blue}{Compare the true statics and our proceedings estimation.}}
        $\mathcal{M} \xleftarrow{} \{j \in [N] \mid  \sum_{n=1}^{N} \hat{\boldsymbol{\mu}}_{n, j}^{(t,m+1)} - \sum_{n=1}^{N} \boldsymbol{\mu}_{n, j}^{(t,m+1)} > 0 \}$\;
        
        $j_{\text{max}} \xleftarrow{} \textbf{argmax}_{j \in \mathcal{M}} \sum_{n=1}^{N} \hat{\boldsymbol{\mu}}_{n, j}^{(t,m+1)} - \sum_{n=1}^{N} \boldsymbol{\mu}_{n, j}^{(t,m+1)}$\;

        $\mathcal{I} \xleftarrow{} \{j \in [N] \mid  \sum_{n=1}^{N} \hat{\boldsymbol{\mu}}_{n, j}^{(t,m+1)} - \sum_{n=1}^{N} \boldsymbol{\mu}_{n, j}^{(t,m+1)} < 0 \}$\;
        
        $j_{\text{min}} \xleftarrow{} \textbf{argmax}_{j \in \mathcal{I}} \sum_{n=1}^{N} \hat{\boldsymbol{\mu}}_{n, j}^{(t,m+1)} - \sum_{n=1}^{N} \boldsymbol{\mu}_{n, j}^{(t,m+1)}$\;

        $N_{j_{\text{max}}}^{(t)} \xleftarrow{} N_{j_{\text{max}}}^{(t)} - m$\;
        
        $N_{j_{\text{min}}}^{(t)} \xleftarrow{} N_{j_{\text{min}}}^{(t)} + m$\;
     }
    \Return $\mathbf{N}^{(t)}: \{N_{j}^{(t)}, \forall j \in [N]\}$
  \caption{Posterior Search}\label{alg2}
\end{algorithm}

\end{document}